\documentclass[10pt, journal,web]{article}

\usepackage{amsthm}

\newtheorem{definition}{Definition}

\usepackage{times}
\usepackage{subcaption}
\usepackage[utf8]{inputenc}
\usepackage{utfsym}
\usepackage{multirow}
\usepackage[normalem]{ulem}
\usepackage{amsmath,amssymb,amsfonts}
\usepackage{algorithmic}
\useunder{\uline}{\ul}{}
\usepackage{url}
\def\BibTeX{{\rm B\kern-.05em{\sc i\kern-.025em b}\kern-.08em
    T\kern-.1667em\lower.7ex\hbox{E}\kern-.125emX}}

\usepackage{graphicx}
\usepackage{authblk}

\usepackage{longtable}
\usepackage{pdflscape}
\usepackage[T1]{fontenc}
\usepackage{bbm}
\usepackage[ruled,vlined]{algorithm2e}
\usepackage{pifont}
\usepackage{xcolor}
\usepackage{hyperref}
\usepackage{booktabs}
\hypersetup{
   colorlinks=true,
    linkcolor=blue,
   citecolor=blue,
    urlcolor=blue
}
\usepackage{makecell}
\usepackage{tabularx}
\usepackage{enumitem}

\makeatletter
\renewcommand\AB@affilsepx{, \protect\Affilfont}
\makeatother

\providecommand{\keywords}[1]{%
  \small
  \textbf{\textit{Keywords---}} #1%
}

\begin{document}

\title{\textbf{Model-Agnostic Fairness Regularization for GNNs with Incomplete Sensitive Information}}

\author[1]{M. Tavassoli Kejani}
\author[2, 3]{F. Dornaika\thanks{Corresponding author}}
\author[1,4]{J. M. Loubes}
\affil[1]{\textit{Institut de Mathématiques de Toulouse}}
\affil[2]{\textit{University of the Basque Country}}
\affil[3]{\textit{IKERBASQUE}}
\affil[4]{\textit{INRIA, Regalia Team}}

\affil[ ]{

\small\texttt{mahdi.tavassoli-kejani@univ-toulouse.fr, fadi.dornaika@ehu.eus, loubes@math.univ-toulouse.fr}}
\date{}

\maketitle

\begin{abstract}
Graph Neural Networks (GNNs) have demonstrated exceptional efficacy in relational learning tasks, including node classification and link prediction. However, their application raises significant fairness concerns, as GNNs can perpetuate and even amplify societal biases against protected groups defined by sensitive attributes such as race or gender. These biases are often inherent in the node features, structural topology, and message-passing mechanisms of the graph itself. A critical limitation of existing fairness-aware GNN methods is their reliance on the strong assumption that sensitive attributes are fully available for all nodes during training—a condition that poses a practical impediment due to privacy concerns and data collection constraints. To address this gap, we propose a novel, model-agnostic fairness regularization framework designed for the realistic scenario where sensitive attributes are only partially available. Our approach formalizes a fairness-aware objective function that integrates both\textit{equal opportunity} and \textit{statistical parity} as differentiable regularization terms. Through a comprehensive empirical evaluation across five real-world benchmark datasets, we demonstrate that the proposed method significantly mitigates bias across key fairness metrics while maintaining competitive node classification performance. Results show that our framework consistently outperforms baseline models in achieving a favorable fairness-accuracy trade-off, with minimal degradation in predictive accuracy. The datasets and source code will be publicly released at \url{https://github.com/mtavassoli/GNN-FC}.

\end{abstract}
 \hspace{10pt}

\section{Introduction}

In recent years, bias mitigation in AI-based algorithms has emerged as a critical concern, as surveyed in \cite{caton2024fairness,mehrabi2021survey}. Beyond ethical considerations, the European Union has introduced mandatory regulations governing algorithms used in high-risk systems, which are defined as those with the potential to cause harm to health, safety, or fundamental rights \footnote{https://artificialintelligenceact.eu/}. These regulations emphasize the need to ensure that such systems do not infringe upon fundamental rights, particularly with respect to preventing discrimination and avoiding unfair biases prohibited by Union or national law.

Addressing bias in the development and evaluation of high-risk AI systems presents significant challenges. In some cases, it may be necessary to process special categories of personal data to enable bias detection and correction, provided this is done in strict accordance with legal conditions and includes appropriate safeguards to protect fundamental rights and privacy. Nevertheless, bias mitigation efforts are often hindered by the unavailability of key sensitive attributes, such as ethnic origin and gender. While the law may permit the collection of such data under certain circumstances to enhance fairness and comply with data protection regulations, these sensitive attributes are frequently scarce and difficult to obtain. For instance, only 14\% of teenage users make their full profiles public on Facebook \cite{madden2013teens}, limiting the availability of necessary information for bias detection.

One class of AI-based models where concerns regarding bias and fairness are increasingly pertinent is the Graph Neural Networks (GNNs) \cite{scarselli2008graph}. GNNs are a subset of deep learning models designed specifically to operate on graph-structured data, enabling the capture of intricate relational and structural information. Due to their capacity for modeling complex dependencies, GNNs have become useful  in various high-stakes applications, including social networks \cite{fan2019graph} and recommendation systems \cite{wu2022graph}. Consequently, addressing issues of bias and fairness in GNNs is of paramount importance. Machine learning algorithms often inherit and can amplify biases present in training data. These biases have been extensively studied in recent years, as discussed in \cite{oneto2020fairness,del2020review,barocas2023fairness} and references therein. For graph neural networks, biases may arise from several sources, including node attributes (referred to as attribute bias in \cite{wang2022improving}), the connections between nodes (structural bias in \cite{wang2022improving}), and the message-passing mechanisms employed in GNNs (potential bias in \cite{wang2022improving}), all of which can result in unfair or biased predictions. For example, in book recommendation systems leveraging GNNs, the model might exhibit a bias towards recommending books by male authors \cite{buyl2020debayes}. Therefore, ensuring the fairness, safety, and reliability of GNN models and their predictions is critical, making this a vital area of ongoing research.

Many methods have been proposed to reduce the effects of bias in traditional machine learning algorithms. In the context of GNNs, existing approaches for bias mitigation remain limited. Several fairness-aware GNN methods have been proposed \cite{dai2021say,dong2022edits,rahman2019fairwalk,agarwal2021towards,Ma_2022,Guo_2023,hu2024migrate,zhang2024learning}, which can be broadly categorized into four groups \cite{chen2024fairness}:
(1) \textbf{pre-processing methods}, which remove bias before applying GNNs, such as \cite{spinelli2021fairdrop}; (2) \textbf{in-training methods}, which address bias during GNN training, like \cite{liu2023generalized}; (3) \textbf{post-processing methods}, which remove bias from the resulting GNN embeddings, as in \cite{kose2023fairness}; and (4) \textbf{hybrid methods}, which combine two or more of the aforementioned techniques, like \cite{kang2020inform}.

While existing methods contribute to improving fairness in AI models, they often come at the cost of reduced accuracy and may even exacerbate it. 
Furthermore, many of these approaches require access to all sensitive attribute labels during the training phase \cite{wu2022graph}, which is not feasible in practical real-world scenarios.

To address these challenges, we propose a novel in-training, model-agnostic approach, \textbf{E}qual \textbf{O}pportunity \textbf{S}tatistical \textbf{P}arity (EOSP), which operates with limited sensitive attribute labels (available only for labeled nodes) during training to enable fair learning in GNNs. To the best of our knowledge, this is the first work to integrate statistical parity and equal opportunity directly as regularization terms for bias mitigation in GNNs. The EOSP method incorporates these two key fairness constraints directly into the training process: a statistical parity term, which ensures similar model behavior across demographic groups, and an equal opportunity term, which guarantees that errors are distributed fairly rather than enforcing identical decisions for everyone.

This approach contrasts with existing adversarial frameworks, such as LAFTR ~\cite{madras2018learning}. While LAFTR reduces the influence of sensitive attributes by training an adversary to predict them, our method directly incorporates fairness penalties. Furthermore, we specifically design and evaluate our approach for the graph learning domain, where the relational structure presents unique challenges for fairness that LAFTR was not designed to address.

The main contributions of the paper are listed below.

\begin{itemize}
    \item \textbf{Model-Agnostic and Composable Framework:} We design a flexible, model-agnostic framework that can be seamlessly integrated into any standard GNN architecture and is composable with other fairness-aware techniques.

    \item \textbf{Effectiveness with Partial Sensitive Attributes:} Our method effectively mitigates bias when sensitive attribute labels are available only for labeled nodes, relaxing the strong and often unrealistic assumption of complete attribute knowledge required by prior methods.

    \item \textbf{Extensive Empirical Validation:} We demonstrate the effectiveness of our approach through comprehensive experiments on five real-world datasets, with particular emphasis on balanced accuracy, a fair performance metric often overlooked in prior fairness literature, alongside standard metrics such as F1-score and AUC.
\end{itemize}

The rest of the paper is structured as follows. We begin by reviewing related work in Section 2 and introducing necessary preliminaries in Section 3. Our proposed methodology is then detailed in Section 4, followed by experimental setup in section 5 and experimental results in Section 6. Finally, we conclude the paper in Section 7.
 
\section{Related Work}\label{sec:relatedwork}

The literature on fairness in machine learning offers various methods for bias mitigation, generally categorized into pre-processing of training data, in-processing with fairness constraints during model optimization, and post-processing of model outputs. Representative approaches include fairness-constrained optimization, latent space disentanglement, optimal transport, and counterfactual reasoning, as discussed in \cite{dwork2012fairness,zafar2017fairness,RisserEtAlJMIV2022,donini2018empirical,gordaliza2019obtaining}.

For graph-based learning, several studies \cite{venkatasubramanian2021fairness,kang2021fair,kang2022algorithmic,khajehnejad2022crosswalk,laclau2021all} explore bias detection and mitigation in Graph Neural Networks (GNNs). According to \cite{Guo_2023}, the main fairness notions considered are group fairness (balanced outcomes across sensitive groups), individual fairness (similar results for similar nodes), and counterfactual fairness (invariance of outcomes under counterfactual scenarios). This work focuses on two main directions of fair GNN: group fairness and counterfactual fairness.

\paragraph{Group Fairness in GNNs}  Group fairness ensures that demographic groups defined by sensitive attributes, such as gender, receive equitable treatment in learning tasks. For example, FairGNN  \cite{dai2021say} employs adversarial learning, where the GNN predicts target labels while an adversarial network simultaneously attempts to infer sensitive attributes from node embeddings. The GNN is trained to minimize task loss while reducing the adversary’s predictive capability, thereby improving fairness. FairVGNN \cite{wang2022improving} enhances fairness by feature channels correlated with sensitive attributes and clamping weights to reduce the influence of sensitive information, supported by adversarial discriminators. FairDrop \cite{spinelli2021fairdrop} proposes an edge-dropout scheme to mitigate homophily and improve fairness in graph representation learning. Fairwalk \cite{rahman2019fairwalk} introduce fair random walk strategies to learn equitable network embeddings, while BIND \cite{dong2023interpreting} proposes the Probabilistic Distribution Disparity (PDD) metric to measure and eliminate bias-contributing nodes. FairSAD \cite{zhu2024fair} separates sensitive attribute information into an independent component with a channel masking mechanism that adaptively identifies and isolates sensitive factors.

More recently, DAB-GNN \cite{lee2025disentangling} introduces a GNN framework that disentangles attribute, structural, and potential biases into separate node embeddings. A bias contrast optimizer ensures that these embeddings remain independent, while a fairness harmonizer aligns subgroup distributions within each bias-specific embedding to reduce the influence of sensitive attributes. FairSR \cite{liu5320563rethinking} integrates structural rebalancing with adversarial learning to ensure equitable outcomes for nodes with varying degrees but similar functionalities in prediction tasks.

\paragraph{Counterfactual Fairness in GNNs} Research on counterfactual fairness in GNNs aims to maintain consistent predictions between factual and counterfactual scenarios. For example, NIFTY \cite{agarwal2021towards} and GEAR \cite{Ma_2022} employ perturbation strategies and GraphVAE-based models to generate realistic counterfactuals and enforce fairness constraints. Despite progress, generating realistic counterfactuals remains challenging. CAF \cite{Guo_2023} addresses this by developing a structural causal model that improves fairness. FairGB \cite{li2024rethinking} tackles unfairness through group balancing using real counterfactual node mixup and a contribution alignment loss. FCLCA \cite{li2024contrastive} proposes a framework to learn counterfactual fairness by addressing graph structure bias. It generates two counterfactual graphs via structural augmentation and employs contrastive learning to maximize consistency between node representations across these augmented graphs. Additionally, FCLCA incorporates adversarial debiasing to further reduce the influence of sensitive attributes, yielding fairer node representations

In this paper, we propose a model-agnostic framework that enforces fairness by incorporating equal opportunity and statistical parity as regularization terms, which can be integrated into standard GNN architectures and combined with other fairness-aware techniques.

\section{Background} \label{sec:background}

\subsection{Notations}
\paragraph{Notations and Problem Setting.} 
We consider a dataset of \( n \) individuals. Each individual \( i \) is associated with a feature vector $\mathbf{x}_i \in \mathbb{R}^d$
and a binary sensitive attribute $\mathrm{s}_i \in \{0, 1\}$ (e.g., gender or race). The goal is to predict a binary target label $\mathrm{y}_i \in \{0, 1\}$. We operate in a semi-supervised setting, where the index set \( \{1, \dots, n\} \) is partitioned into a labeled set \( \mathcal{L} = \{1, \dots, \ell\} \) and an unlabeled set \( \mathcal{U} = \{\ell + 1, \dots, n\} \), with \( n = \ell + u \).
Both target labels $\mathrm{y}_i$ and sensitive attributes $\mathrm{s}_i$ are assumed to be observed only for individuals in the labeled set \( \mathcal{L} \).
\paragraph{Graph Data} We model the data as a graph \( \mathcal{G} = (\mathcal{V}, \mathcal{E}) \), where the node set \( \mathcal{V} \) (with \( |\mathcal{V}| = n \)) represents the individuals, and the edge set \( \mathcal{E} \subseteq \mathcal{V} \times \mathcal{V} \) represents relationships between them. The graph structure is encoded by an adjacency matrix \( \mathbf{A} \in \{0, 1\}^{n \times n} \), where \( \mathbf{A}_{ij} = 1 \) if an edge exists between nodes \( i \) and \( j \), and \( 0 \) otherwise. For standard graph-theoretic definitions, we refer the reader to \cite{bondy2008graph}. Although node features $\mathbf{x_i}$ may be used to construct the graph, in this work we assume a given graph structure and a binary sensitive attribute for clarity. The proposed framework can, however, be extended to weighted graphs and non-binary sensitive attributes.

\paragraph{Objective} 
Our objective is to learn a fair classifier that accurately predicts target labels $\mathrm{Y}$ while minimizing dependence on sensitive attributes $\mathrm{S}$. We employ an encoder-classifier framework where:
\begin{itemize}
    \item A GNN encoder \( f_{\theta}: \mathbb{R}^{n \times d} \times \{0,1\}^{n \times n} \rightarrow \mathbb{R}^{n \times d'} \) maps the input features and graph structure to a set of node representations \( \mathbf{H} = \{\mathbf{h}_1, \dots, \mathbf{h}_n\} \in \mathbb{R}^{n \times d'} \), where \(\mathbf{h}_i\) is the node representation for individual \(i\).
   \item A classifier \( g_{\phi}: \mathbb{R}^{n \times d'} \rightarrow (0,1)^{n \times 1} \) maps the node representations to a vector of predicted probabilities \(\mathrm{P} = (\mathrm{p}_1, \dots, \mathrm{p}_n)\), where \( \mathrm{p}_i \in (0,1) \) is the predicted probability for individual \(i\).
   \item Final predictions are obtained through thresholding: $\mathrm{\hat{Y}} = \mathbb{I}[\mathrm{P} > 0.5]$, where $\mathrm{\hat{y}}_i$ is the predicted label for individual \(i\).
\end{itemize}

We seek representations that are predictive of target labels while invariant to sensitive attributes.

\subsection{Graph Neural Networks}
Graph Neural Networks (GNNs) learn node-level representations through message passing, where each node aggregates feature information from its neighbors. GNN architectures including GCN \cite{kipf2016semi}, GIN \cite{xu2019powerful}, GraphSAGE \cite{hamilton2017inductive}, and GAT \cite{liu2022graph} implement this approach. For semi-supervised label propagation, a two-layer GCN model is formulated as:

\begin{equation}
    \mathbf{H} = \sigma\left(\hat{\mathbf{A}} \ \sigma\left(\hat{\mathbf{A}} \mathbf{X} \mathbf{W}^{(0)}\right) \mathbf{W}^{(1)}\right)
\end{equation}
where $\mathbf{X}$ is a feature matrix, $ \hat{\mathbf{A}} =\mathbf{D}^{-1/2}\mathbf{A}\mathbf{D}^{-1/2}$ is the normalized adjacency matrix, ${\mathrm{D}}_{ii} = \sum_j \mathrm{A}_{ij}$ is the degree matrix, $\sigma(\cdot)$ is the ReLU activation function, $\mathbf{W}^{(0)} \in \mathbb{R}^{d \times h}$ is the input-to-hidden weight matrix, and $\mathbf{W}^{(1)} \in \mathbb{R}^{h \times d'}$ is the hidden-to-hidden weight matrix.

For binary classification, the model outputs probabilities as:

\begin{equation}
    \mathrm{P} = g_{\phi}\left({\mathbf{H}}\right)
\end{equation}

The base model is is trained by minimizing the cross-entropy loss over labeled nodes:
\begin{equation}
    \mathcal{L}_{\text{pred}} = -\frac{1}{|\mathcal{L}|} \sum_{i \in \mathcal{L}} \left[\mathrm{y}_i \log(\mathrm{p}_i) + (1-\mathrm{y}_i) \log(1-\mathrm{p}_i)\right]
    \label{eq:pred}
\end{equation}

where $\mathrm{p}_i \in (0,1)$ is the predicted probability and $\mathrm{y}_i \in \{0,1\}$ is the corresponding class label.
 
This objective alone may lead to biased predictions.

\subsection{Fairness in Graph Neural Networks}

Evaluating fairness in Graph Neural Networks (GNNs) is a critical research concern that has been explored from multiple fundamental perspectives. Chen et al.~\cite{chen2023fairness} categorize fairness evaluation metrics for GNNs into four levels: \textit{prediction-level}, \textit{graph-level}, \textit{neighborhood-level}, and \textit{embedding-level} metrics. In this work, we focus on \textit{prediction-level fairness}, which assesses how model outputs are influenced by sensitive attributes. Within this context, two key fairness notions are commonly considered: \textit{statistical parity} and \textit{equal opportunity}, defined as follows:

\paragraph{Statistical Parity (SP)} requires predictions to be statistically independent of sensitive attributes, ensuring equal selection rates across groups:
\begin{equation}
    \Delta_{\mathrm{SP}} = \left| P(\mathrm{\hat{y}} = 1 \mid \mathrm{s} = 1) - P(\mathrm{\hat{y}} = 1 \mid \mathrm{s} = 0) \right|
\end{equation}

\paragraph{Equal Opportunity (EO)} requires equal true positive rates across sensitive groups, ensuring qualified individuals have equal chances of positive outcomes:
\begin{equation}
    \Delta_{\mathrm{EO}} = \left| P(\mathrm{\hat{y}} = 1 \mid \mathrm{y} = 1, \mathrm{s} = 1) - P(\mathrm{\hat{y}} = 1 \mid \mathrm{y} = 1, \mathrm{s} = 0) \right|
\end{equation}

Lower values of both \( \Delta_{\mathrm{SP}} \) and \( \Delta_{\mathrm{EO}} \) indicate better fairness. In the literature, the two metrics mentioned above are commonly used to assess the fairness of trained GNNs.  

\section{Fairness-Aware Regularization Framework}

Our proposed method, Equal Opportunity Statistical Parity (EOSP), integrates fairness directly into GNN training through differentiable regularization terms.   While fairness is typically evaluated on test data, we employ the labeled training set as a practical approximation to guide the model toward fair representations.
In other words, our approach explicitly integrates fairness metrics into the end-to-end training of a given GNN, considering them as differentiable losses.

\subsection{Statistical Parity Regularizer}

Statistical Parity requires that predictions are independent of sensitive attributes. While the evaluation metric operates on discrete predictions:

\begin{equation}
    \Delta_{SP} = \left| P(\mathrm{\hat{y}} = 1 \mid \mathrm{s} = 0) - P(\mathrm{\hat{y}} = 1 \mid \mathrm{s} = 1) \right|
    \label{eq:evalSP}
\end{equation}

we propose a differentiable approximation for training and used it as a loss function:

\begin{equation}
    \mathcal{L}_{\text{SP}} = \left| \frac{1}{|\mathcal{D}_1|} \sum_{i \in \mathcal{D}_1} \mathrm{p}_i - \frac{1}{|\mathcal{D}_0|} \sum_{j \in \mathcal{D}_0} \mathrm{p}_j \right|
    \label{eq:fairSP}
\end{equation}
where \(\mathcal{D}_1 = \{ i \in \mathcal{L} : \mathrm{s}_i = 1 \}\) and \(\mathcal{D}_0 = \{ j \in \mathcal{L} : \mathrm{s}_j = 0 \}\) represent the sensitive groups in the labeled training set.

This loss is differentiable because it uses continuous probabilities $\mathrm{p}_x$ instead of discrete predictions $\mathrm{\hat{y}}_x$, enabling gradient-based optimization.

\subsection{Equal Opportunity Regularizer}

Equal Opportunity requires equal true positive rates across sensitive groups. While the evaluation metric uses discrete predictions:

\begin{equation}
    \Delta_{EO} = \left| P(\mathrm{\hat{y}} = 1 \mid \mathrm{y} = 1, \mathrm{s} = 0) - P(\mathrm{\hat{y}} = 1 \mid \mathrm{y} = 1, \mathrm{s} = 1) \right|
    \label{eq:evalEO}
\end{equation}

we define a differentiable approximation for training and used it as a loss function:

\begin{equation}
    \mathcal{L}_{\text{EO}} = \left| \frac{1}{|\mathcal{P}_1|} \sum_{i \in \mathcal{P}_1} \mathrm{p}_i - \frac{1}{|\mathcal{P}_0|} \sum_{j \in \mathcal{P}_0} \mathrm{p}_j \right|
    \label{eq:fairEO}
\end{equation}
where \(\mathcal{P}_1 = \{ i \in \mathcal{L} : \mathrm{y}_i = 1, \mathrm{s}_i = 1 \}\) and \(\mathcal{P}_0 = \{ j \in \mathcal{L} : \mathrm{y}_j = 1, \mathrm{s}_j = 0 \}\).

This formulation is differentiable as it operates on probability scores $\mathrm{p}_x$ rather than binary decisions, allowing seamless integration with gradient-based learning.

\subsection{Final Objective Function}
The Graph Neural Network (GNN) model introduced in the previous section optimizes predictive accuracy but does not account for its potential dependency on sensitive attributes \(s\). To learn a model that is both accurate and fair, we propose regularizing the standard training objective with a composite fairness loss. Consequently, the model is trained to minimize the following global loss function:
\begin{equation}
    \mathcal{L}_{\text{total}} = \mathcal{L}_{\text{pred}} +  \mathcal{L}_{\text{fairness}}
    \label{GCN}
\end{equation}

Here, \(\mathcal{L}_{\text{pred}}\) is the standard cross-entropy loss (Eq.~(\ref{eq:pred})) for the node classification task. The term \(\mathcal{L}_{\text{fairness}}\) is our proposed fairness regularizer, defined as a weighted sum of two distinct fairness criteria:
\begin{equation}
    \mathcal{L}_{\text{fairness}} =  \alpha \, \mathcal{L}_{\text{eo}}(g_{\phi}(f_{\theta}(\mathbf{X}, \mathbf{A})) , \mathrm{S}, \mathrm{Y}) + \beta \, \mathcal{L}_{\text{sp}}(g_{\phi}(f_{\theta}(\mathbf{X}, \mathbf{A})), \mathrm{S})
    \label{fairness}
\end{equation}

In this formulation, \(\mathcal{L}_{\text{EO}}\) enforces equal opportunity and \(\mathcal{L}_{\text{SP}}\) enforces statistical parity. The hyperparameters \(\alpha\) and \(\beta\) provide explicit control over the trade-off between these two fairness notions. Although these objectives are based on different statistical criteria, we reproduce in the Appendix the proof from \cite{defrance2025maximal} showing that they can be combined.

 We employ Bayesian hyperparameter optimization \cite{akiba2019optuna} to tune \(\alpha\) and \(\beta\). A key advantage of this framework is its model-agnostic nature. For any generic GNN model \(f_{\theta}\), regardless of whether fairness is incorporated into its design, with a native loss function \(\mathcal{L}_{\text{GNN}}\), it can be  trained by minimizing:
\begin{equation}
    \mathcal{L}_{\text{fair}} = \mathcal{L}_{\text{GNN}} + \mathcal{L}_{\text{fairness}}
    \label{eq:total_loss_generic}
\end{equation}

This approach provides a flexible and direct method for integrating fairness constraints into existing GNN architectures and other fairness-aware techniques. The following section empirically demonstrates this integration in several standard GNN and fairness-aware models. To ensure reproducibility and facilitate further research, we provide our implementation at: \url{https://github.com/mtavassoli/GNN-FC}.

\section{Experimental Setup}
This section presents the comprehensive methodology used to evaluate the model-agnostic EOSP fairness constraint. We detail the datasets employed, the compared methods, training settings, evaluation metrics, and provide in-depth implementation details.

\subsection{Datasets}
We evaluated the proposed method on five publicly available real-world tabular datasets, commonly used in graph-based fairness research. A summary of each dataset is provided below:

\begin{enumerate}
    \item \textbf{German \cite{asuncion2007uci}:}  
    This dataset consists of 1000 individual clients from a German bank. The main objective is to evaluate the credit risk level for each individual, with a focus on the sensitive attribute "gender". In this dataset, the nodes represent individuals, and the links are based on significant similarities in their bank account details.
    
    \item \textbf{Bail \cite{jordan2015effect}:}  
    This dataset contains records of 18,876 individuals granted bail by US state courts between 1990 and 2009. The primary goal is to classify whether a defendant is granted bail or not, focusing on the sensitive attribute "race". In this dataset, the nodes represent individuals, and the links are based on similarities in their demographics and prior criminal histories.
    
    \item \textbf{Credit Defaulter \cite{yeh2009comparisons}:}  
    This dataset includes 30,000 individual credit card holders. The main objective is to predict whether an individual will default on credit card payments, with an emphasis on the sensitive attribute "age". In this dataset, the nodes represent individuals, and the links are based on high similarity in payment information.

    \item \textbf{NBA \cite{dai2021say}:}
    The dataset includes 400 NBA players with statistics from the 2016–2017 season, along with information such as nationality, age and salary. The players act as nodes in a graph, linked based on their Twitter connections. The main task is to predict whether a player’s salary is above or below the league median, with nationality (U.S. or non-U.S.) used as a sensitive attribute.
    
    \item \textbf{Pokec-n \cite{takac2012data}:}  
    This dataset is based on a popular online social network in Slovakia, similar to Facebook, and includes 66,569 individual users. The main objective is to predict a user's working field, with an emphasis on the sensitive attribute "region". In this dataset, the nodes represent users, and the links are based on friendship relations between them.
\end{enumerate}

\subsection{Compared Methods}
We evaluate our model-agnostic EOSP method by integrating it with both standard GNN architectures (GCN \cite{kipf2016semi}, GraphSAGE \cite{hamilton2017inductive}) and state-of-the-art fairness-aware methods (FairGNN \cite{dai2021say}, FairVGNN \cite{wang2022improving}, CAF \cite{Guo_2023}, FairGB \cite{li2024rethinking}, DAB-GCN \cite{lee2025disentangling}). For each model, we compare performance with and without EOSP to assess its impact on both accuracy and fairness.

\subsection{Training Settings}
For all methods, we tune the learning rate (\(lr \in \{10^{-2}, 10^{-3}\}\)), hidden size (\(h \in \{16, 32\}\)), dropout rate (\(dr = 0.2\)), and model depth (\(\text{layers} \in \{1, 2, 3\}\)). Hyperparameters for baseline methods follow their original publications. Train/validation/test splits follow \cite{dong2023fairness}, with labeled samples set to 100 (German), 100 (Bail), 6000 (Credit), 100 (NBA), and 500 (Pokec-n). Validation and test sets each comprise 25\% of the data, allocated independently. All methods during training use labeled sensitive attributes equal to the training set size, except for CAF, which uses all available sensitive attributes.

Each experiment is run five times with different data splits and trained for 100 epochs. For the EOSP method, the hyperparameters \(\alpha\) and \(\beta\) are tuned via Bayesian optimization using Optuna~\cite{akiba2019optuna}. The search space for both \(\alpha\) and \(\beta\) includes the values $
\{0.01, 0.02, \ldots, 0.1, 0.2, 0.3, \ldots, \\
\qquad 1, 2, 5\}$.  
 Optimization is performed with 15 trials, and the best combination is chosen based on validation set performance. Hyperparameters of baseline methods are kept fixed to ensure a fair comparison.

All experiments run on nodes with dual Intel Xeon Gold 6136 processors (24 cores total, 3 GHz) and 192 GB RAM, using Python 3.9.18.

\subsection{Evaluation Metrics}
We evaluate both classification performance and prediction-level fairness. For classification, we use balanced accuracy (BACC), area under the ROC curve (AUC), and F1 score. For fairness, we employ statistical parity difference \(\Delta_{SP}\) (Eq.~(\ref{eq:evalSP})) and equal opportunity difference \(\Delta_{EO}\) (Eq.~(\ref{eq:evalEO})) metrics, where all metrics are expressed as percentages.

To select the best epoch during training, we compute the following hybrid score evaluated on the validation data:
\begin{equation}
\text{Score} = \text{BACC} + \frac{1}{2} \left[ (100 - \Delta_{EO}) + (100 - \Delta_{SP}) \right]
\label{eq:hybridscore}
\end{equation}

\subsection{Implementation of the Model-Agnostic EOSP Fairness Constraint Using GNN-FC}

The proposed EOSP fairness constraint is implemented as a modular component in the open-source \texttt{GNN-FC} library\footnote{\url{https://github.com/mtavassoli/GNN-FC}}, which provides compatibility with standard GNN architectures (e.g., GCN) and fairness-aware frameworks (e.g., FairGNN). As outlined in Algorithm~\ref{eosp-integration}, the EOSP module integrates as an auxiliary objective during model training, ensuring full compatibility with PyTorch’s automatic differentiation framework.

\begin{algorithm}
\caption{Integration of the EOSP Fairness Constraint}
\label{eosp-integration}
\begin{algorithmic}[1]
\REQUIRE class labels $Y_{\mathcal{L}}$, sensitive attribute labels $S_{\mathcal{L}}$, Hyperparameters ($\alpha$,$\beta$)
\REQUIRE 
\FOR{iteration $= 1$ to $T$}
    \STATE Compute model loss $\mathcal{L}_{\text{GNN}}$
    \STATE Compute fairness loss (EOSP) $\mathcal{L}_{\text{fairness}}$ (Eq.~(\ref{fairness}))
    \STATE Compute total fair loss $\mathcal{L}_{\text{total}} = \mathcal{L}_{\text{GNN}} + \mathcal{L}_{\text{fairness}}$ (Eq.~(\ref{eq:total_loss_generic}))
     \STATE  Update the model by back-propagation
\ENDFOR
\end{algorithmic}
\end{algorithm}

\section{Experimental Results}
In this section, we aim to thoroughly evaluate our proposed model-agnostic EOSP framework. To this end, we address the following research questions:

\begin{enumerate} 
    \item \textbf{RQ1}: How do our proposed fairness losses affect the fairness and performance of existing GNNs?
    \item \textbf{RQ2}: How sensitive is our method to its hyperparameters (\(\alpha\) and \(\beta\))?
    \item \textbf{RQ3}: How does limited labeled sensitive attribute availability affect fairness-aware graph learning?
    \item \textbf{RQ4}: What is the computational overhead of our fairness method?
\end{enumerate}
\subsection{RQ1: Performance Comparison}
To evaluate the effectiveness of the proposed model-agnostic EOSP framework, we conducted a comparative analysis of classification performance and fairness outcomes across a range of models, both with and without the EOSP fairness constraint. Experiments were performed on the German, Bail, Credit Defaulter, NBA, and Pokec-n datasets. The results, including the means and standard deviations over five runs with different random splits of train/validation/test, are reported as percentages in Tables~\ref{tab:German}, \ref{tab:Bail}, \ref{tab:CreditDefaulter}, \ref{tab:NBA}, and \ref{tab:PokecN}.

\begin{table}[!h]
    \centering
    \caption{Evaluation of Node Classification and Fairness Performance on the German Dataset. The best result between methods with and without EOSP is bold.}\label{tab:German}
     \resizebox{0.96\textwidth}{!}{%
    \begin{tabular}{lrrrrr}
        \toprule
        \textbf{Method} & 
        \textbf{BACC($\uparrow$)} & 
        \textbf{AUC($\uparrow$)} & \textbf{F1($\uparrow$)} & \boldmath$\Delta_{SP}(\downarrow)$ & \boldmath$\Delta_{EO}(\downarrow)$ \\
        \midrule
        \textbf{GCN \cite{kipf2016semi}} &  59.70( 3.79) & \textbf{63.29(5.00)} & \textbf{69.25 (5.43)} & 5.58(7.64) & 5.36(6.10) \\
        \textbf{GCN-EOSP (Ours)} & \textbf{60.02(5.25)} & 63.21(5.32) & 63.89(8.75) & \textbf{1.98(1.81)} & \textbf{1.86(1.04)} \\
        \midrule
        \textbf{SAGE \cite{hamilton2017inductive}} & 60.34(4.71) & 65.37(5.23) & 53.82(23.20) & \textbf{6.98(6.58)} & 11.23(6.52) \\
        \textbf{SAGE-EOSP (Ours)} & \textbf{62.90(1.66)} & \textbf{67.68(3.12)} & \textbf{68.78(3.47)} & 11.63(3.50) & \textbf{10.37(3.67)} \\
        \midrule
        \textbf{FairGNN  \cite{dai2021say}} & 61.01(4.56) & 64.63(3.75) & 67.18(3.05) & 10.04(5.72) & 8.52(6.51) \\
        \textbf{FairGNN-EOSP (Ours)} & \textbf{61.35(4.03)} & \textbf{65.18(3.52)} & \textbf{68.79(3.95)} & \textbf{6.90(5.27)} &  \textbf{4.94(2.31)} \\
        \midrule
        \textbf{FairVGNN \cite{wang2022improving}} & \textbf{61.50(2.59)} & \textbf{65.51(2.83)} & \textbf{71.37(4.43)} & 5.28(4.06) & \textbf{4.16(2.02)} \\
        \textbf{FairVGNN-EOSP (Ours)} & 60.19(4.79) & 65.36(3.09) & 68.00(7.85) & \textbf{4.04(1.29)} & 4.46(3.46)\\
        \midrule
        \textbf{CAF \cite{Guo_2023}} & 61.68(4.02) & 66.08(4.19) & \textbf{69.27(6.58)} & 6.25(5.66) & 9.42(5.83) \\
        \textbf{CAF-EOSP (Ours)} &  \textbf{62.13(5.55)} & \textbf{66.11(4.84)} & 67.22(6.39) & \textbf{4.13(1.46)} & \textbf{4.19(2.33)} \\
        \midrule
        \textbf{Fair-GB \cite{li2024rethinking}} & \textbf{61.62(3.51)} & 66.16(5.44) & 65.77(7.95) & 7.78(6.83) & 8.57(7.31) \\
        \textbf{FairGB-EOSP (Ours)} & 61.10(3.52) & \textbf{66.44(4.69)} & \textbf{68.35(8.49)} & \textbf{7.24(4.73)} & \textbf{5.47(3.23)} \\
        \midrule
        \textbf{DAB-GCN\cite{lee2025disentangling}} & \textbf{60.72(3.19)} & 65.02(2.87) & 67.01(3.69) & 5.93(3.77) & 5.77(4.31) \\
        \textbf{DAB-GCN-EOSP (Ours)} & 60.59(1.56) & \textbf{65.90(2.44)} & \textbf{67.01(2.40)} & \textbf{5.13(1.54)} & \textbf{5.62(2.73)} \\
        \midrule
        \textbf{Average Improvement} & \textbf{+0.24} & \textbf{+0.55} & \textbf{+1.34} & \textbf{+0.83} & \textbf{+2.16}
        \\
        \bottomrule
    \end{tabular}
    }
\end{table}
On the German dataset (Table~\ref{tab:German}), applying EOSP to GCN, FairGNN, CAF, FairGB, and DAB-GCN yields consistent fairness improvements while maintaining comparable classification accuracy. This confirms that EOSP enhances fairness without compromising predictive utility. GCN-EOSP achieves the strongest fairness performance among all models, with $\Delta_{SP}=1.98$ and $\Delta_{EO}=1.86$, representing substantial improvements over both vanilla GNNs and existing fairness-aware baselines. For classification performance, SAGE-EOSP attains the highest predictive scores, achieving a BACC of 62.90, AUC of 67.68, and F1-score of 68.78.
\begin{table}[!h]
    \centering
    \caption{Evaluation of Node Classification and Fairness Performance on the Bail Dataset. The best result between methods with and without EOSP is bold.}\label{tab:Bail}
     \resizebox{0.96\textwidth}{!}{%
    \begin{tabular}{lrrrrr}
        \toprule
        \textbf{Method} & 
        \textbf{BACC($\uparrow$)} & 
        \textbf{AUC($\uparrow$)} & \textbf{F1($\uparrow$)} & \boldmath$\Delta_{SP}(\downarrow)$ & \boldmath$\Delta_{EO}(\downarrow)$ \\
        \midrule
        \textbf{GCN \cite{kipf2016semi}} & 91.07 (1.07) & 93.94 (1.41) & 89.21 (1.26) & 7.18 (1.10) & 1.57 (0.37) \\
        \textbf{GCN-EOSP (Ours)} & \textbf{91.90 (0.98)} & \textbf{94.67 (1.30)} & \textbf{90.12 (1.20)} & \textbf{6.89 (0.82)} & \textbf{1.47 (0.80)} \\
        \midrule
        \textbf{SAGE \cite{hamilton2017inductive}} & \textbf{94.87 (0.93)} & \textbf{97.72 (0.50)} & \textbf{93.82 (0.94)} & 7.49 (1.24) & 1.38 (0.93) \\
        \textbf{SAGE-EOSP (Ours)} & 94.84 (1.00) & 97.63 (0.51) & 93.71 (1.39) & \textbf{6.91 (1.04)} & \textbf{1.33 (0.98)} \\
        \midrule
        \textbf{FairGNN  \cite{dai2021say}} & 90.37 (0.97) & \textbf{91.94 (1.48)} & 86.15 (1.48) & 6.89 (1.18) & 2.82 (1.20) \\
        \textbf{FairGNN-EOSP (Ours)} & \textbf{90.77 (0.54)} & 91.32 (1.96) & \textbf{86.26 (2.11)} & \textbf{6.86 (0.74)} & \textbf{2.24 (1.18)} \\
        \midrule
        \textbf{FairVGNN \cite{wang2022improving}} & \textbf{91.85 (1.04)} & \textbf{95.68 (1.11)} & 89.65 (1.26) & 7.10 (1.12) & \textbf{0.59 (0.20)} \\
        \textbf{FairVGNN-EOSP (Ours)} & 91.67 (0.60) & 95.04 (0.96) & \textbf{89.76 (0.83)} & \textbf{6.16 (1.27)} & 1.24 (0.81) \\
        \midrule
        \textbf{CAF \cite{Guo_2023}} & 94.43 (1.44) & 97.16 (0.85) & 93.08 (1.87) & 6.88 (1.40) & 1.20 (0.80) \\
        \textbf{CAF-EOSP (Ours)} & \textbf{95.10 (0.99)} & \textbf{97.67 (0.63)} & \textbf{93.62 (1.30)} & \textbf{5.93 (1.40)} & \textbf{0.53 (0.27)} \\
        \midrule
        \textbf{FairGB \cite{li2024rethinking}} & 93.06 (0.86) & 96.35 (0.82) & 91.75 (0.85) & 6.68 (1.49) & 1.35 (1.05) \\
        \textbf{FairGB-EOSP (Ours)} & \textbf{94.50 (0.52)} & \textbf{97.46 (0.48)} & \textbf{93.18 (0.75)} & \textbf{5.82 (1.51)} & \textbf{1.05 (1.97)} \\
        \midrule
        \textbf{DAB-GCN \cite{lee2025disentangling}} & 71.55 (6.19) & \textbf{79.89 (5.04)} & 63.18 (9.92) & 11.67 (10.86) & 10.40 (7.86) \\
        \textbf{DAB-GCN-EOSP (Ours)} & \textbf{72.85 (4.99)} & 79.85 (4.79) & \textbf{66.25 (5.76)} & \textbf{4.28 (4.38)} & \textbf{4.67 (4.92)} \\
         \midrule
        \textbf{Average Improvement} & \textbf{+0.64} & \textbf{+0.14}	& \textbf{+0.84}	& \textbf{+1.56}	& \textbf{+0.96}
        \\
        \bottomrule
    \end{tabular}
    }
\end{table}

On the Bail dataset (Table~\ref{tab:Bail}), EOSP consistently improves both fairness and predictive performance across GCN, SAGE, FairGNN, CAF, FairGB, and DAB-GCN. These results demonstrate the effectiveness of EOSP for both vanilla GNNs and existing fairness-aware baselines. Among the models, CAF-EOSP achieves the strongest fairness gains, with \(\Delta_{SP} = 5.93\) and \(\Delta_{EO} = 0.53\), substantially reducing disparities compared to all other models. It also attains the best classification performance, achieving the highest balanced accuracy (95.10), AUC (97.67), and F1-score (93.62). These findings highlight EOSP's ability to achieve a superior balance between fairness and predictive accuracy.

\begin{table}[!h]
    \centering
    \caption{Evaluation of Node Classification and Fairness Performance on the Credit Dataset. The best result between methods with and without EOSP is bold.}\label{tab:CreditDefaulter}
     \resizebox{0.96\textwidth}{!}{%
    \begin{tabular}{lrrrrr}
        \toprule
        \textbf{Method} & 
        \textbf{BACC($\uparrow$)} & 
        \textbf{AUC($\uparrow$)} & \textbf{F1($\uparrow$)} & \boldmath$\Delta_{SP}(\downarrow)$ & \boldmath$\Delta_{EO}(\downarrow)$ \\
        \midrule
        \textbf{GCN \cite{kipf2016semi}} & 64.15 (0.37) & \textbf{69.16 (0.58)} & \textbf{81.34 (0.36)} & 9.61 (2.07) & 8.39 (2.26) \\
        \textbf{GCN-EOSP (Ours)} & \textbf{64.29 (0.33)} & 69.03 (0.40) & 79.85 (0.70) & \textbf{1.54 (0.71)} & \textbf{1.48 (1.32)} \\
        \midrule
        \textbf{SAGE \cite{hamilton2017inductive}} & 69.62 (0.29) & \textbf{75.57 (0.44)} & 82.91 (2.99) & 7.44 (3.04) & 6.09 (3.00) \\
        \textbf{SAGE-EOSP (Ours)} & \textbf{69.95 (0.56)} & 75.42 (0.60) & \textbf{83.62 (1.02)} & \textbf{1.61 (1.06)} & \textbf{1.47 (0.78)} \\
        \midrule
        \textbf{FairGNN  \cite{dai2021say}} & 63.73 (0.48) & \textbf{70.27 (0.20)} & 78.16 (0.54) & 11.13 (3.58) & 10.12 (3.99) \\
        \textbf{FairGNN-EOSP (Ours)} & \textbf{64.37 (0.23)} & 70.00 (0.29) & \textbf{79.46 (1.61)} & \textbf{1.20 (0.42)} & \textbf{0.71 (0.83)} \\
        \midrule
        \textbf{FairVGNN \cite{wang2022improving}} & 64.97 (1.12) & 69.56 (0.92) & \textbf{85.53 (2.03)} & 6.46 (2.93) & 5.58 (1.83) \\
        \textbf{FairVGNN-EOSP (Ours)} & \textbf{65.79 (0.28)} & \textbf{69.87 (0.38)} & 79.05 (2.13) & \textbf{2.07 (0.95)} & \textbf{2.34 (1.37)} \\
        \midrule
        \textbf{CAF \cite{Guo_2023}} & 69.43 (1.14) & \textbf{75.37 (0.39)} & 80.61 (4.74) & 5.48 (3.22) & 4.66 (2.22) \\
        \textbf{CAF-EOSP (Ours)} & \textbf{69.57 (0.48)} & 75.02 (0.34) & \textbf{80.91 (1.28)} & \textbf{2.31 (1.45)} & \textbf{1.93 (0.66)} \\
        \midrule
        \textbf{FairGB \cite{li2024rethinking}} & 69.63 (0.44) & 74.54 (0.13) & 82.14 (2.95) & 1.55 (1.05) & 1.63 (0.85) \\
        \textbf{FairGB-EOSP (Ours)} & \textbf{69.93 (0.25)} & \textbf{74.74 (0.40)} & \textbf{84.59 (0.62)} & \textbf{0.93 (0.49)} & \textbf{1.12 (0.75)} \\
        \midrule
        \textbf{DAB-GCN \cite{lee2025disentangling}} & 69.11 (1.08) & 72.25 (0.62) & 80.90 (2.57) & 6.57 (3.17) & 5.04 (3.27) \\
        \textbf{DAB-GCN-EOSP (Ours)} & \textbf{69.58 (0.33)} & \textbf{73.80 (0.34)} & \textbf{81.95 (0.95)} & \textbf{1.76 (1.42)} & \textbf{2.03 (0.46)} \\
         \midrule
        \textbf{Average Improvement} & \textbf{+0.40}	& \textbf{+0.31}	& -0.31	& \textbf{+5.28}	& \textbf{4.36}
        \\
        \bottomrule
    \end{tabular}
    }
\end{table}

On the Credit dataset (Table~\ref{tab:CreditDefaulter}), applying EOSP improves fairness across all vanilla GNNs and existing fairness-aware baselines, reducing $\Delta_{SP}$ and $\Delta_{EO}$ by over 80\% in some cases while maintaining comparable classification accuracy. This confirms that EOSP enables fairness improvements without compromising predictive utility. FairGB-EOSP achieves the strongest fairness performance among all models, with $\Delta_{SP}=0.93$ and $\Delta_{EO}=1.12$, representing substantial improvements over both vanilla GNNs and existing fairness-aware baselines. For classification performance, SAGE-EOSP and FairGB-EOSP attain the highest predictive scores, achieving BACC of 69.95 and 69.93 respectively with superior F1 scores.

\begin{table}[!h]
    \centering
    \caption{Evaluation of Node Classification and Fairness Performance on the NBA Dataset. The best result between methods with and without EOSP is bold.}\label{tab:NBA}
     \resizebox{0.96\textwidth}{!}{%
    \begin{tabular}{lccccc}
        \toprule
        \textbf{Method} & 
        \textbf{BACC($\uparrow$)} & 
        \textbf{AUC($\uparrow$)} & \textbf{F1($\uparrow$)} & \boldmath$\Delta_{SP}(\downarrow)$ & \boldmath$\Delta_{EO}(\downarrow)$ \\
        \midrule
        \textbf{GCN \cite{kipf2016semi}} & 65.51 (2.64) & \textbf{74.36 (2.31)} & 66.83 (3.38) & \textbf{7.22 (4.67)} & 15.27 (9.22) \\
        \textbf{GCN-EOSP (Ours)} & \textbf{67.85 (3.83)} & 74.19 (2.75) & \textbf{67.72 (2.25)} & 8.47 (4.64) & \textbf{14.89 (3.21)} \\
        \midrule
        \textbf{SAGE \cite{hamilton2017inductive}} & 62.03 (3.27) & 69.90 (5.92) & 59.27 (14.74) & 9.62 (6.90) & 17.46 (16.61) \\
        \textbf{SAGE-EOSP (Ours)} & \textbf{69.11 (2.20)} & \textbf{77.41 (2.05)} & \textbf{69.34 (3.31)} & \textbf{8.16 (3.92)} & \textbf{9.98 (9.20)} \\
        \midrule
        \textbf{FairGNN  \cite{dai2021say}} & 68.25 (9.03) & 79.62 (4.28) & 66.29 (6.46) & 7.74 (7.44) & 16.40 (14.71) \\
        \textbf{FairGNN-EOSP (Ours)} & \textbf{70.79 (3.97)} & \textbf{81.79 (2.48)} & \textbf{66.87 (10.61)} & \textbf{6.97 (4.04)} & \textbf{10.61 (10.13)} \\
        \midrule
        \textbf{FairVGNN \cite{wang2022improving}} & 65.81 (2.48) & 73.08 (2.41) & 65.10 (6.98) & \textbf{6.71 (4.48)} & 16.15 (6.88) \\
        \textbf{FairVGNN-EOSP (Ours)} & \textbf{70.09 (2.18)} & \textbf{74.76 (3.36)} & \textbf{71.07 (1.70)} & 9.30 (4.32) & \textbf{11.06 (6.13)} \\
        \midrule
        \textbf{CAF \cite{Guo_2023}} & 66.68 (4.58) & 71.60 (4.85) & \textbf{70.74 (2.59)} & 8.60 (4.32) & \textbf{7.45 (6.09)} \\
        \textbf{CAF-EOSP (Ours)} & \textbf{66.69 (6.20)} & \textbf{73.56 (3.33)} & 70.38 (3.77) & \textbf{4.30 (4.86)} & 7.47 (8.56) \\
        \midrule
        \textbf{FairGB \cite{li2024rethinking}} & \textbf{68.22 (3.23)} & \textbf{74.00 (2.55)} & \textbf{71.64 (2.66)} & 13.85 (9.15) & \textbf{10.34 (8.55)} \\
        \textbf{FairGB-EOSP (Ours)} & 67.53 (2.35) & 72.05 (2.69) & 69.30 (2.75) & \textbf{11.67 (5.64)} & 13.34 (8.47) \\
        \midrule
        \textbf{DAB-GCN \cite{lee2025disentangling}} & 74.07 (3.80) & 80.51 (4.27) & 76.51 (2.17) & 12.30 (6.13) & 12.03 (7.00) \\
        \textbf{DAB-GCN-EOSP (Ours)} & \textbf{74.34 (4.96)} & \textbf{80.99 (2.65)} & \textbf{76.52 (3.58)} & \textbf{10.37 (4.75)} & \textbf{11.10 (8.15)} \\
        \midrule
        \textbf{Average Improvement} & \textbf{+2.27} & \textbf{+2.24}	& \textbf{+1.99} & \textbf{+0.97} & \textbf{+2.43}
        \\
        \bottomrule
    \end{tabular}
    }
\end{table}

On the NBA dataset (Table~\ref{tab:NBA}), EOSP enhances both fairness and predictive performance across GCN, SAGE, and DAB-GCN models. CAF-EOSP delivers the strongest fairness improvements, achieving \(\Delta_{SP} = 4.30\) and \(\Delta_{EO} = 7.47\), substantially reducing disparities compared to baseline methods. In terms of classification performance, FairGNN-EOSP achieves the highest AUC (81.79), while DAB-GCN-EOSP attains the best balanced accuracy (74.34) and F1-score (76.52). These results demonstrate EOSP's effectiveness in balancing fairness and predictive utility.

\begin{table}[!h]
    \centering
    \caption{Evaluation of Node Classification and Fairness Performance on the Pokec-n Dataset. The best result between methods with and without EOSP is bold.}\label{tab:PokecN}
     \resizebox{0.86\textwidth}{!}{%
    \begin{tabular}{lrrrrr}
        \toprule
        \textbf{Method} & 
        \textbf{BACC($\uparrow$)} & 
        \textbf{AUC($\uparrow$)} & \textbf{F1($\uparrow$)} & \boldmath$\Delta_{SP}(\downarrow)$ & \boldmath$\Delta_{EO}(\downarrow)$ \\
        \midrule
        \textbf{GCN \cite{kipf2016semi}} & 64.65 (1.24) & 69.21 (1.40) & 61.46 (3.26) & 1.85 (1.01) & 2.37 (1.58) \\
        \textbf{GCN-EOSP (Ours)} & \textbf{66.30 (0.72)} & \textbf{70.26 (0.70)} & \textbf{62.76 (1.03)} & \textbf{1.45 (0.75)} & \textbf{2.15 (1.50)} \\
        \midrule
        \textbf{SAGE \cite{hamilton2017inductive}} & 64.36 (2.44) & 69.44 (2.31) & \textbf{60.50} (4.27) & \textbf{1.56 (0.96)} & \textbf{2.24 (1.95)} \\
        \textbf{SAGE-EOSP (Ours)} & \textbf{67.04 (0.81)} & \textbf{71.86 (0.69)} & 60.47 (3.92) & 1.55 (1.30) & 3.03 (1.90) \\
        \midrule
        \textbf{FairGNN  \cite{dai2021say}} & 68.34 (0.96) & 72.84 (0.98) & 63.98 (1.12) & 4.18 (3.56) & 5.10 (4.51) \\
        \textbf{FairGNN-EOSP (Ours)} & \textbf{68.45 (1.31)} & \textbf{73.19 (0.90)} & \textbf{64.87 (1.81)} & \textbf{2.90 (1.17)} & \textbf{3.24 (2.40)} \\
        \midrule
        \textbf{FairVGNN} & 66.38 (1.21) & \textbf{70.17 (0.45)} & 58.91 (3.91) & 4.23 (1.42) & 5.44 (2.51) \\
        \textbf{FairVGNN-EOSP (Ours)} & 66.09 (0.78) & 69.71 (0.75) & \textbf{59.89 (1.54)} & \textbf{3.03 (1.39)} & \textbf{4.08 (2.04)} \\
        \midrule
         \textbf{CAF \cite{Guo_2023}} & 62.43 (1.65) & 65.76 (1.80) & \textbf{61.46 (3.22)} & 1.64 (0.53) & \textbf{1.78 (1.81)} \\
        \textbf{CAF-EOSP (Ours)} & \textbf{63.56 (2.32)} & \textbf{67.91 (2.19)} & 61.38 (2.61) & \textbf{1.19 (0.41)} & 2.65 (1.14) \\
        \midrule
        \textbf{FairGB \cite{li2024rethinking}} & 65.08 (2.45) & 70.11 (1.60) & \textbf{61.15 (5.91)} & 2.13 (0.60) & \textbf{2.76 (1.01)} \\
        \textbf{FairGB-EOSP (Ours)} & \textbf{67.14 (1.00)} & \textbf{71.05 (1.49)} & 60.65 (2.76) & \textbf{1.40 (0.66)} & 4.15 (1.17) \\
        \midrule
        \textbf{DAB-GCN \cite{lee2025disentangling}} & 62.03 (1.91) & 66.85 (1.96) & \textbf{60.64 (3.60)} & \textbf{3.54 (2.73)} & 4.15 (2.22) \\
        \textbf{DAB-GCN-EOSP (Ours)
        } & \textbf{62.47 (1.62)} & \textbf{67.09 (1.93)} & 59.97 (3.24) & 3.68 (2.59) & \textbf{4.02 (2.71)} \\
         \midrule
        \textbf{Average Improvement} & \textbf{+1.10}	& \textbf{+0.81} & \textbf{+0.36} & \textbf{+0.56} & \textbf{+0.08}

        \\
        \bottomrule
    \end{tabular}
`    }
\end{table}

On the Pokec-n dataset (Table~\ref{tab:PokecN}), applying EOSP to GCN, FairGNN, and FairVGNN yields consistent fairness improvements while maintaining or enhancing predictive accuracy. CAF-EOSP achieves the strongest statistical parity performance with \(\Delta_{SP} = 1.19\), while CAF attains the best equal opportunity value of \(\Delta_{EO} = 1.78\). For classification performance, FairGNN-EOSP reaches the highest balanced accuracy (68.45) and AUC (73.19). These results demonstrate that EOSP generally provides balanced enhancements across both fairness and utility objectives on this dataset.

The key comparisons evaluate each baseline model against its EOSP-enhanced version across five datasets. Results demonstrate that EOSP generally improves fairness metrics (\(\Delta_{SP}\) and \(\Delta_{EO}\)) while maintaining classification performance (BACC, AUC, F1) for most models. However, in a few cases, EOSP does not uniformly improve fairness metrics; for example, with SAGE-EOSP on the German dataset, \(\Delta_{SP}\) increased from 6.98 to 11.63 compared to the baseline, while \(\Delta_{EO}\) improved modestly from 11.23 to 10.37, alongside increased predictive performance. Although it performs better than the baseline, this raises the question of whether EOSP may not be equally effective for all models or if other factors contribute to its suboptimal performance in this case.

\begin{figure}[h!]
    \centering
        \includegraphics[width=0.6\textwidth]{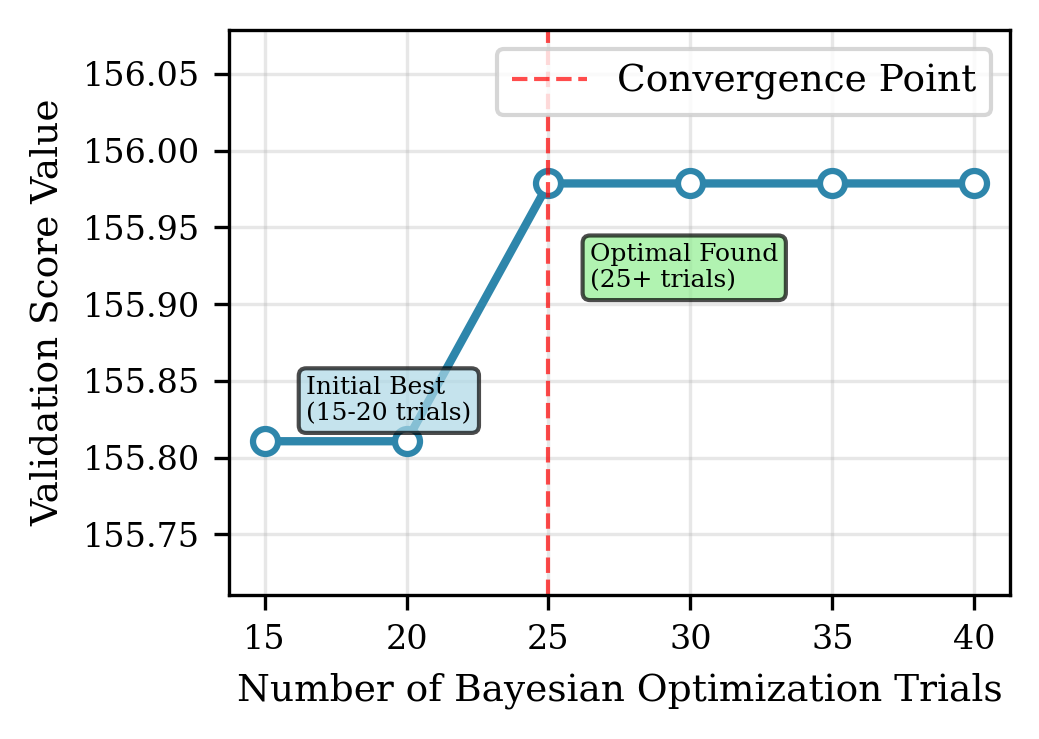}
        \caption{Bayesian Optimization Convergence}
      \label{fig:convergence1} 
\end{figure}

Due to computational constraints, we employ Bayesian hyperparameter optimization (15 trials vs. 400 exhaustive) with hyperparameters $\alpha$ and $\beta$ selected using the validation score defined in Eq.~(\ref{eq:hybridscore}). Figure~\ref{fig:convergence1} shows the convergence trajectory for SAGE-EOSP on the German dataset using a single train/validation/test split, revealing an initial plateau (trials 15-20) followed by optimal discovery at trial 25 and subsequent stability. This demonstrates efficient near-optimal solution finding (16× faster than exhaustive search) and shows that additional trials enable overcoming initial limitations.

\begin{figure}[h!]
    \centering
    \begin{subfigure}[b]{0.45\textwidth}
        \centering
        \includegraphics[width=\textwidth]{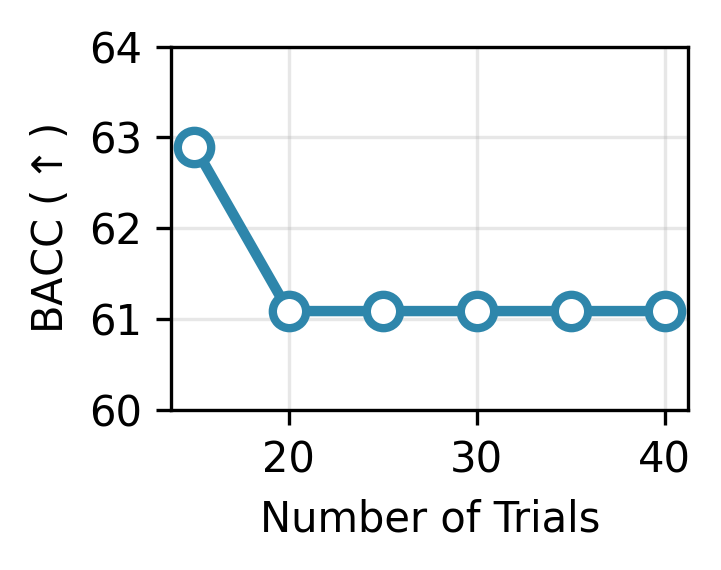}
        \caption{BACC ($\uparrow$)}
        \label{fig:eo_effect}
    \end{subfigure}
    \hfill
    \begin{subfigure}[b]{0.45\textwidth}
        \centering
        \includegraphics[width=\textwidth]{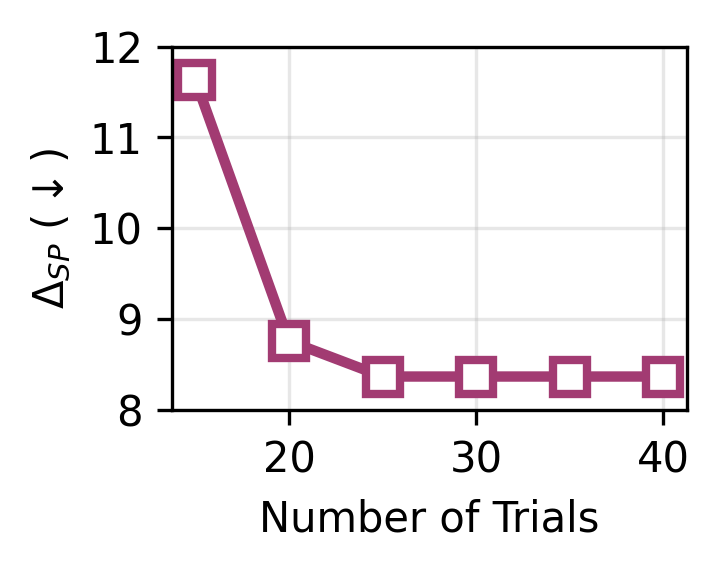}
        \caption{$\Delta_{SP}$ ($\downarrow$)}
        \label{fig:sp_effect}
    \end{subfigure}
     \hfill
    \begin{subfigure}[b]{0.45\textwidth}
        \centering
        \includegraphics[width=\textwidth]{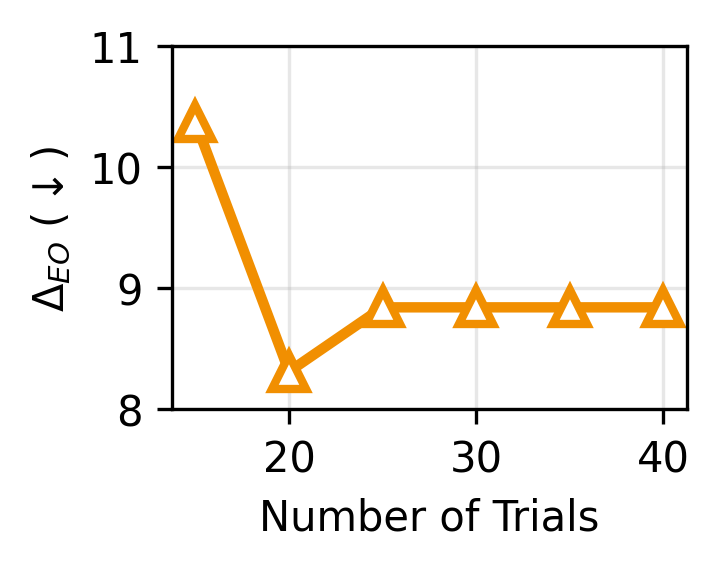}
        \caption{$\Delta_{EO}$ ($\downarrow$)}
        \label{fig:sp_effect}
    \end{subfigure}
    \caption{Bayesian optimization for SAGE-EOSP on the German dataset: (a) Balanced accuracy decreases slightly but stabilizes after 25 trials; (b) Statistical parity ($\Delta_{SP}$) shows significant improvement, decreasing from 11.63 to 8.37; (c) Equal opportunity ($\Delta_{EO}$) converges to stable values. All metrics reach convergence within 25 trials.}
      \label{fig:convergence} 
\end{figure}

Our analysis reveals that extended optimization enables SAGE-EOSP on the German dataset to achieve superior fairness-accuracy-stability trade-offs. While the initial 15-trial configuration exhibited increased $\Delta_{SP}$, optimization over 25 trials yields balanced improvements: $\Delta_{SP}$ improves from 11.63 to 8.37 (baseline: 6.93) and $\Delta_{EO}$ from 11.23 to 8.84, while maintaining competitive accuracy (BACC: 61.09 vs. 60.34 baseline) and dramatically enhancing stability (F1 std: 23.20→3.55). Figure~\ref{fig:convergence} confirms consistent improvements across trials, demonstrating EOSP's robustness.

\begin{figure}[h!]
    \centering
    \begin{subfigure}[b]{0.45\textwidth}
        \centering
        \includegraphics[width=\textwidth]{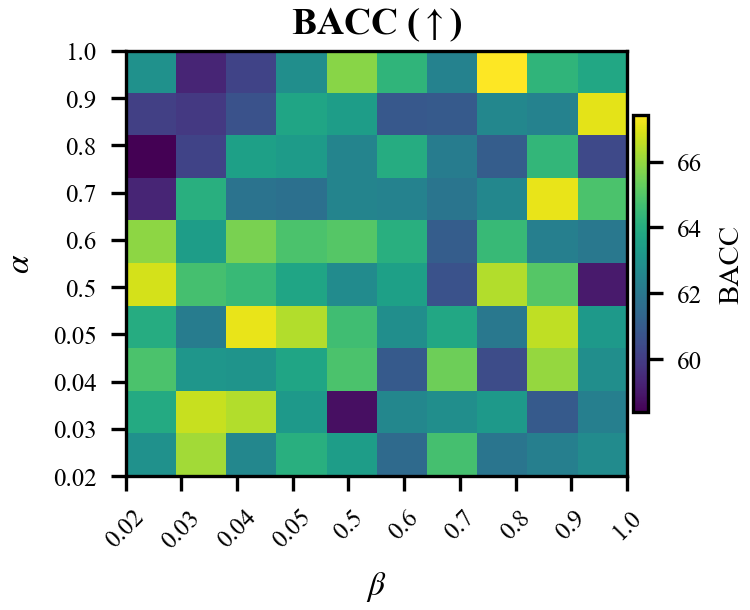}
        \caption{{Performance} ($BACC$ ($\uparrow$))}
        \label{fig:GermanBACC}
    \end{subfigure}
    
    \begin{subfigure}[b]{0.45\textwidth}
        \centering
        \includegraphics[width=\textwidth]{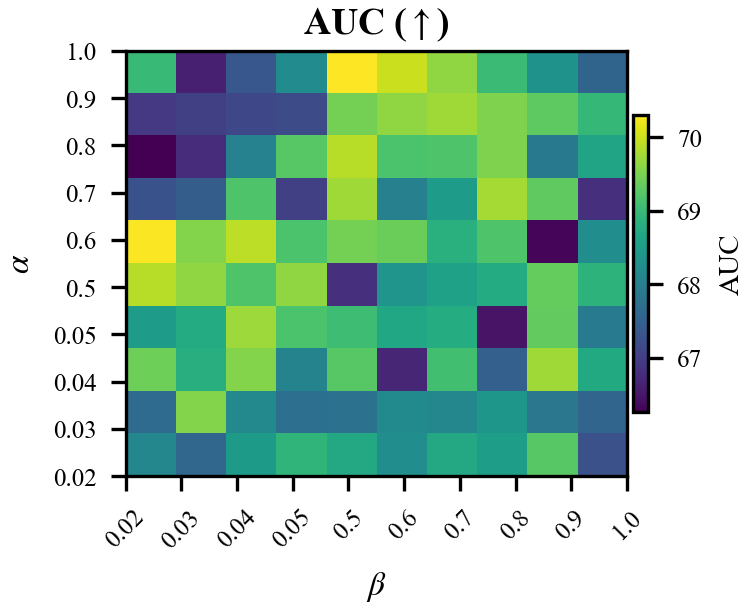}
        \caption{Performance (AUC ($\uparrow$))}
        \label{fig:GermanAUC}
    \end{subfigure}
    \hfill
    \begin{subfigure}[b]{0.45\textwidth}
        \centering
        \includegraphics[width=\textwidth]{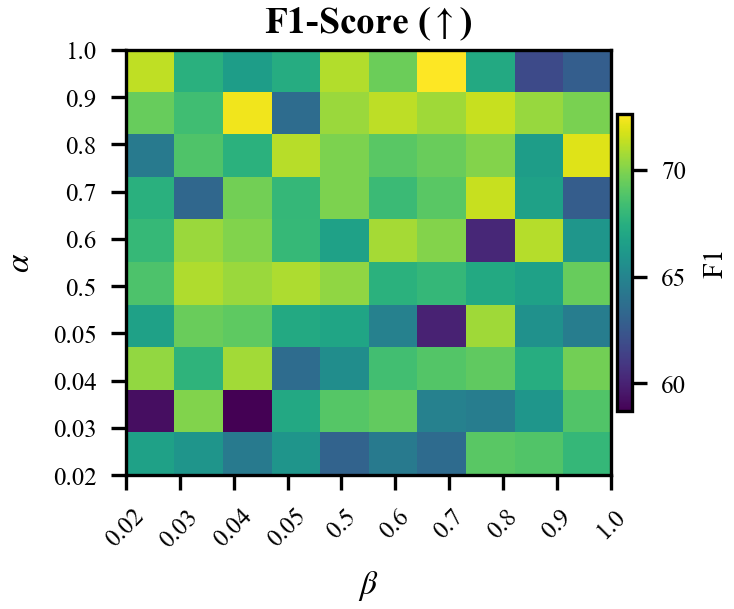}
        \caption{Performance (F1 ($\uparrow$))}
        \label{fig:GermanF1}
    \end{subfigure}

    \begin{subfigure}[b]{0.45\textwidth}
        \centering
        \includegraphics[width=\textwidth]{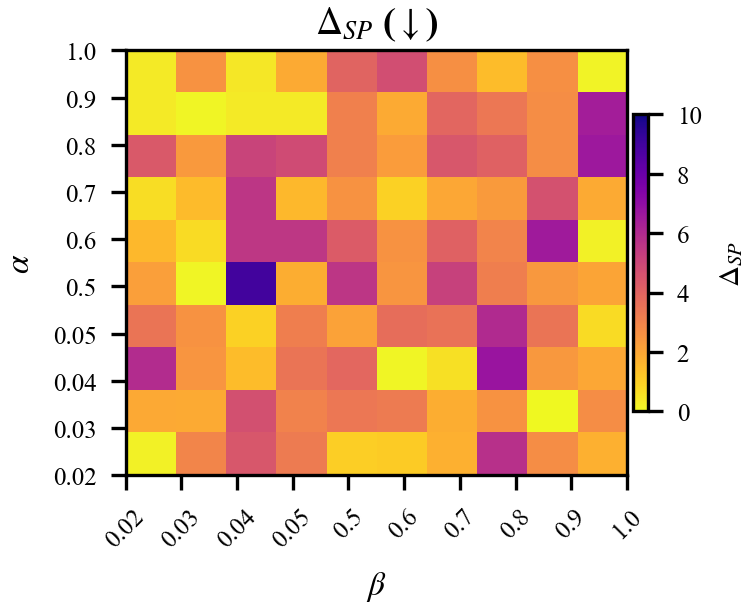}
        \caption{Fairness ($\Delta_{SP}$ ($\downarrow$))}
        \label{fig:GermanSP}
    \end{subfigure}
    \hfill
    \begin{subfigure}[b]{0.45\textwidth}
        \centering
        \includegraphics[width=\textwidth]{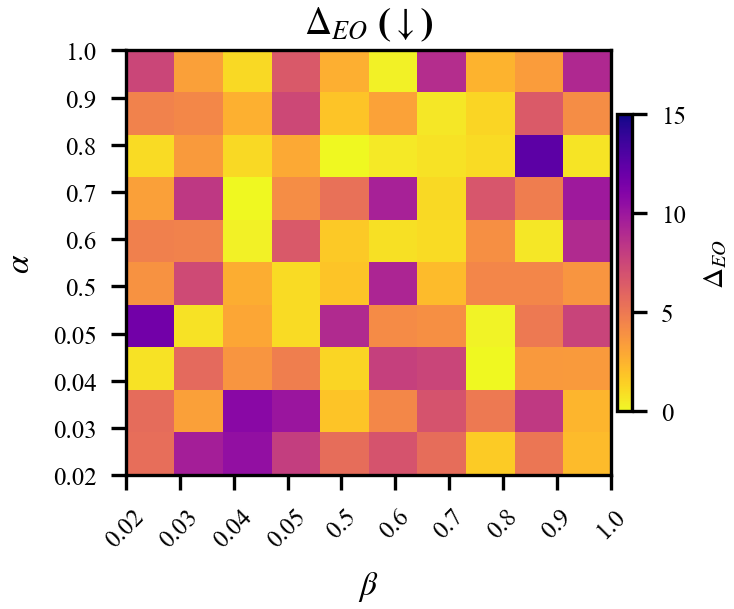}
        \caption{Fairness ($\Delta_{EO}$ ($\downarrow$))}
        \label{fig:GermanEO}
    \end{subfigure}

    \caption{Hyper-parameter analysis on the German dataset, evaluating the effect of varying \( \alpha \)  and \( \beta \) on predictive performance and fairness metrics.}
      \label{fig:germancombinedfigures} 
\end{figure}

\subsection{RQ2: Hyper-Parameter Analysis}
Our proposed EOSP framework introduces two key hyperparameters: \(\alpha\), which controls the influence of the equal opportunity loss, and \(\beta\), which controls the statistical parity loss contribution. \textbf{While both relate to fairness objectives, they impact model accuracy differently, creating a complex trade-off space that requires careful balancing.}

For optimal model configuration in our main experiments, we utilized Bayesian hyperparameter optimization to efficiently explore a discretized two-dimensional parameter space (defined as a rectangular grid) and to identify configurations that effectively balance the competing objectives.

However, to systematically characterize the sensitivity and interaction effects of these parameters \textbf{on both fairness and accuracy}, we conducted a comprehensive grid search with both hyperparameters varying across \{0.02, 0.03, 0.04, 0.05, 0.5, 0.6, \ldots, 1.0\}, covering both fine-grained low values and broader high values to capture their effects.

Figure~\ref{fig:germancombinedfigures} presents the hyperparameter ablation study on the German dataset, illustrating how variations in \(\alpha\) and \(\beta\) affect both predictive performance (BACC, F1, AUC) and fairness metrics (\(\Delta_{EO}\), \(\Delta_{SP}\)). The grid analysis reveals several important patterns: the configuration \(\alpha=0.02\), \(\beta=0.8\) achieves the highest balanced accuracy (67.43) with strong fairness properties (\(\Delta_{SP}=1.41\), \(\Delta_{EO}=2.51\)). However, the optimal fairness metrics occur at distinct operating points: perfect statistical parity (\(\Delta_{SP}=0.00\)) is achieved at \(\alpha=0.9\), \(\beta=0.9\) (BACC = 60.95, AUC = 67.86, F1 = 66.00, \(\Delta_{EO}=8.17\)), while optimal equal opportunity (\(\Delta_{EO}=0.03\)) occurs at \(\alpha=0.04\), \(\beta=0.5\) (BACC = 62.48, AUC = 69.87, F1 = 69.87, \(\Delta_{SP}=3.12\)). This demonstrates that no single configuration simultaneously optimizes all objectives, highlighting the challenge in identifying optimal hyperparameters for balancing the accuracy-fairness trade-off.

This complexity motivates our two-stage approach. First, we define a comprehensive validation metric to jointly evaluate accuracy–fairness trade-offs. This score is given in Eq. (\ref{eq:hybridscore}). Second, we employ Bayesian hyperparameter optimization to efficiently explore the high-dimensional parameter space and identify configurations that effectively balance these competing objectives. This approach is essential, as performing an exhaustive search across the entire hyperparameter space is computationally infeasible.

\begin{figure}[h]
    \centering
    \begin{subfigure}[b]{0.45\textwidth}
        \centering
        \includegraphics[width=\textwidth]{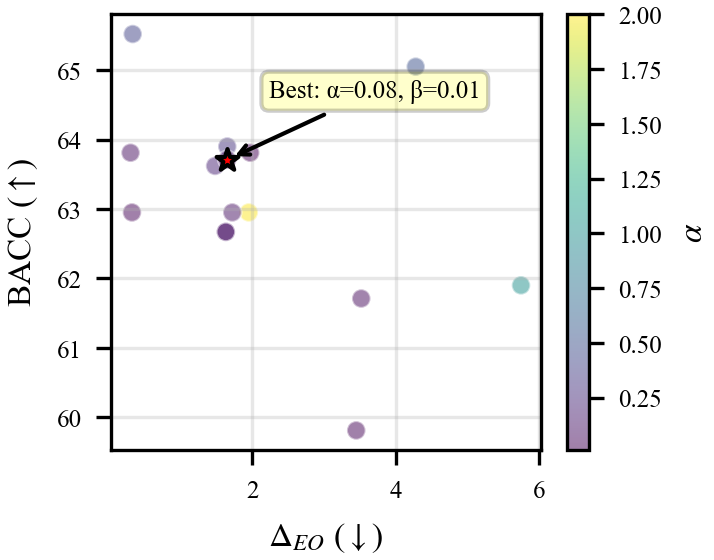}
        \caption{Effect of ($\alpha$)}
        \label{fig:eo_effect}
    \end{subfigure}
    \hfill
    \begin{subfigure}[b]{0.45\textwidth}
        \centering
        \includegraphics[width=\textwidth]{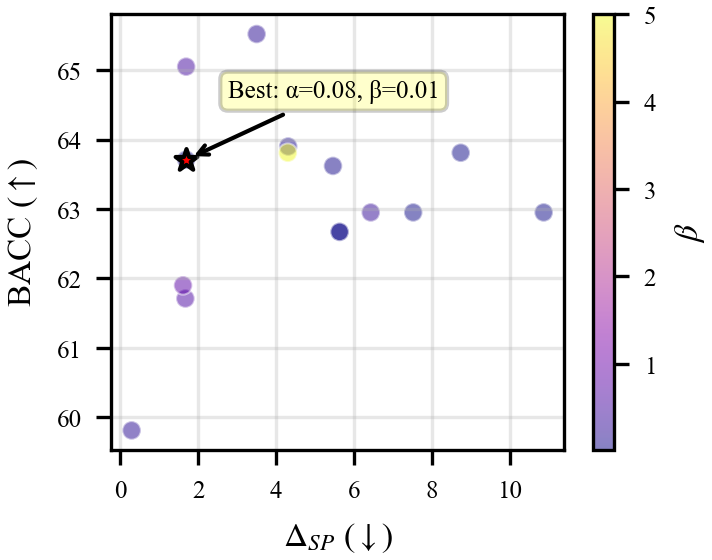}
        \caption{Effect of ($\beta$)}
        \label{fig:sp_effect}
    \end{subfigure}
\caption{Hyperparameter analysis on the German dataset using Bayesian optimization, evaluating the effects of the fairness coefficients $\alpha$ (equal opportunity) and $\beta$ (statistical parity) on the accuracy-fairness trade-off.}
      \label{fig:effect} 
\end{figure}

Figure~\ref{fig:effect} presents the results of our Bayesian hyperparameter optimization of the hybrid score in Eq. (\ref{eq:hybridscore}), showing the relationship between fairness metrics ($\Delta_{EO}$, $\Delta_{SP}$) and balanced accuracy across different $\alpha$ and $\beta$ configurations. The analysis reveals that Bayesian optimization successfully navigated the complex fairness-accuracy trade-off space, identifying an optimal configuration ($\alpha=0.08$, $\beta=0.01$) where minimal fairness regularization achieves the best balance. This suggests that gentle fairness constraints are sufficient to guide the model toward equitable solutions without compromising predictive performance.

The success of Bayesian optimization in this context underscores its value for fairness-aware hyperparameter tuning, efficiently identifying subtle configurations that would be challenging to discover through exhaustive grid search or manual experimentation.

\begin{figure}[h!]
    \centering
    \begin{subfigure}[b]{0.45\textwidth}
        \centering
        \includegraphics[width=\textwidth]{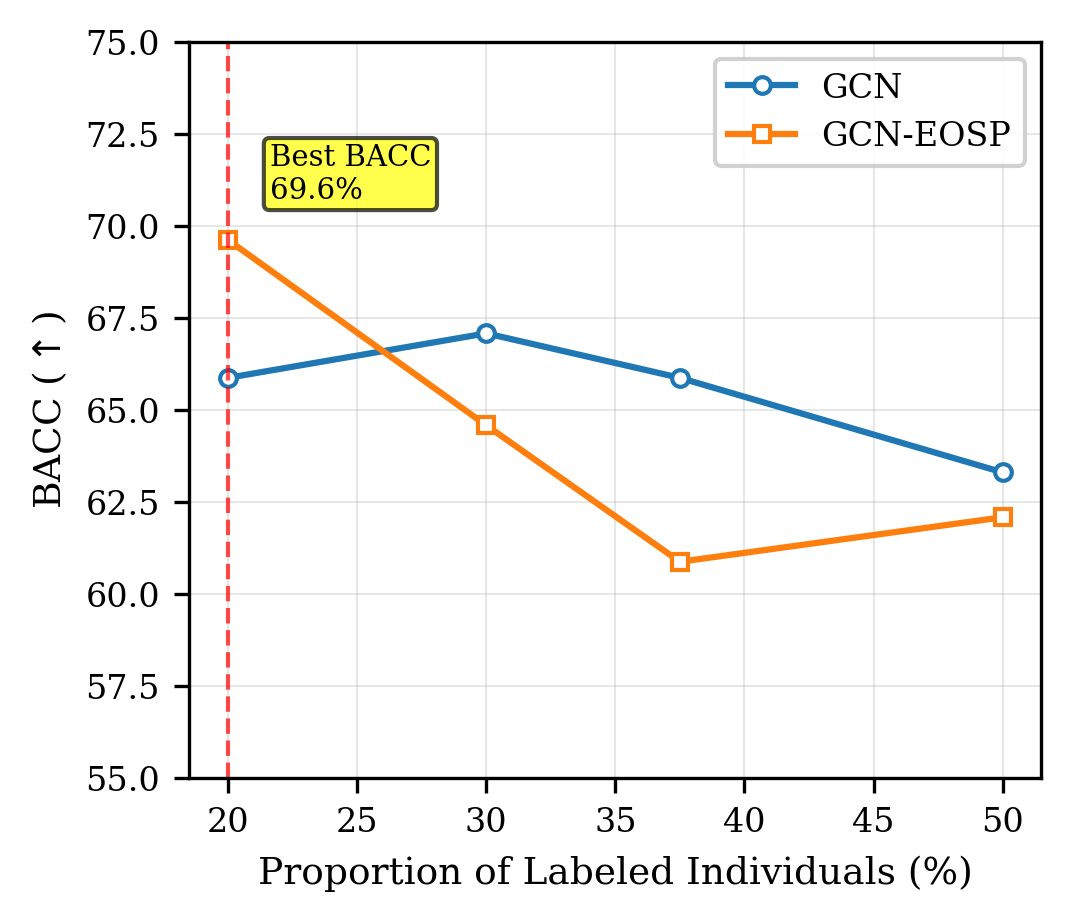}
        \label{fig:bacc_small}
    \end{subfigure}
    \hfill
    \begin{subfigure}[b]{0.45\textwidth}
        \centering
        \includegraphics[width=\textwidth]{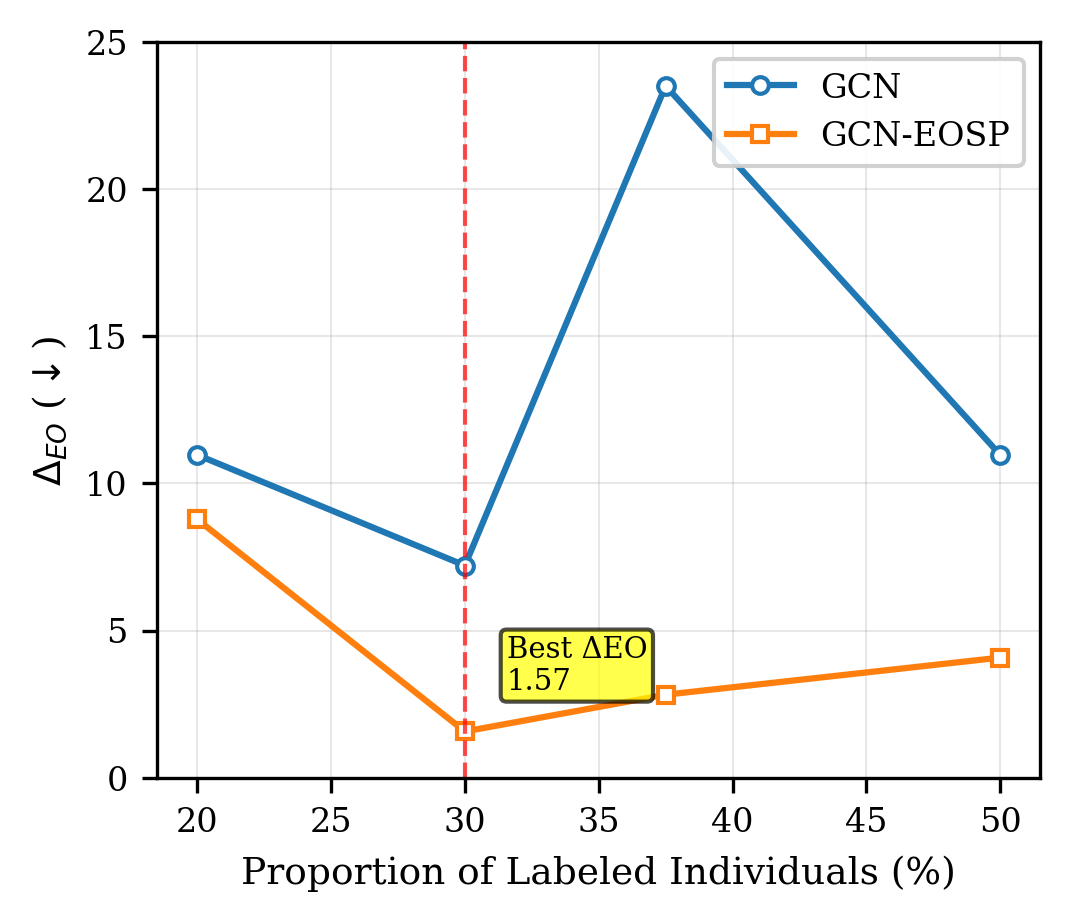}
        \label{fig:sp_effect}
    \end{subfigure}
    \hfill
    \begin{subfigure}[b]{0.45\textwidth}
        \centering
        \includegraphics[width=\textwidth]{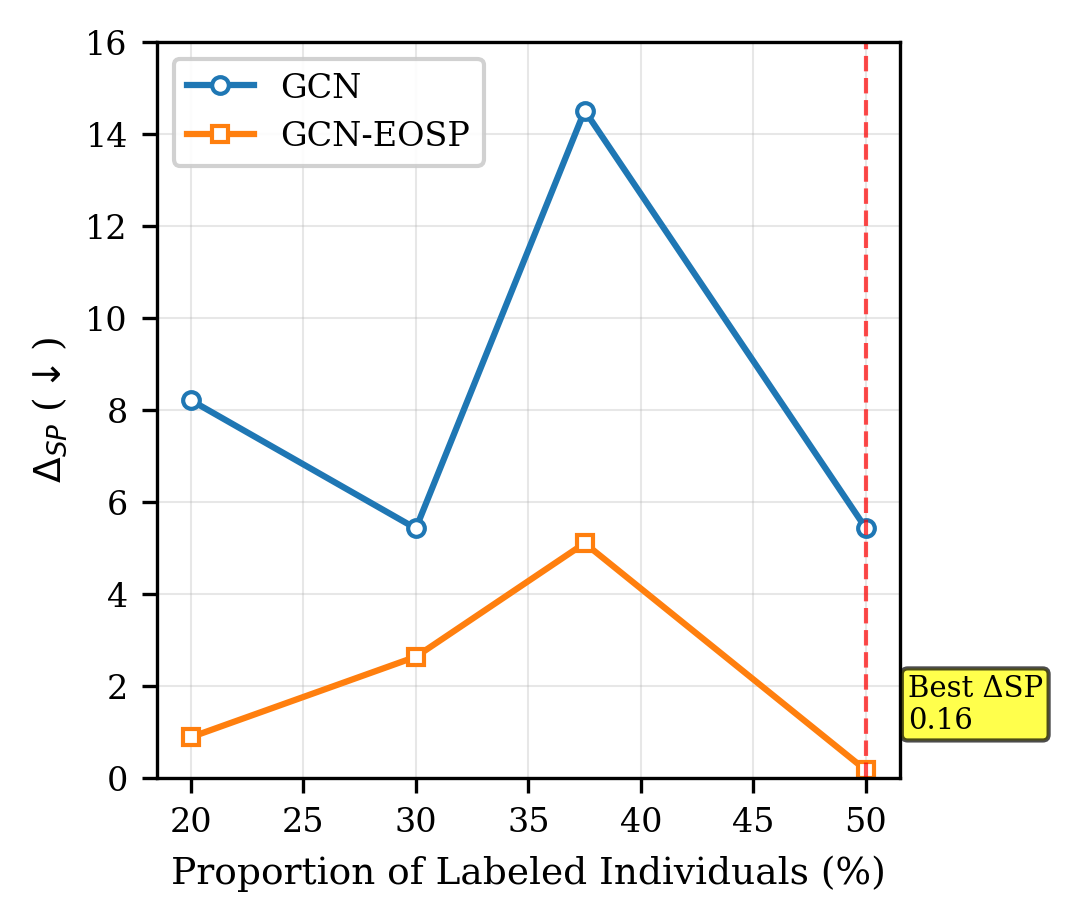}
        \label{fig:sp_effect}
    \end{subfigure}

    \caption{Impact of labeled sensitive attribute proportion on GCN vs GCN-EOSP performance across three metrics.}
      \label{fig:labeld} 
\end{figure}

\subsection{RQ3: Impact of Labeled Sensitive Attribute Availability}

To answer RQ3, we investigate how the number of labeled sensitive attributes affects our proposed EOSP framework. We evaluate GCN and GCN-EOSP on the NBA dataset, varying the proportion of labeled sensitive attributes across \(\{20\%, 30\%, 37.5\%, 50\%\}\) while maintaining an equivalent number of labeled individuals. Recall that for each labeled sample, both the class label and the corresponding sensitive attribute are available.

Figure~\ref{fig:labeld} reveals that GCN-EOSP consistently outperforms vanilla GCN on fairness metrics. Notably, EOSP achieves \textbf{near-perfect statistical parity} (\(\Delta_{SP}=0.16\)) at 50\% labeled attributes, representing a 97\% improvement over GCN, while maintaining competitive accuracy. The optimal configuration varies by metric: balanced accuracy peaks at 20\% labels (BACC = 69.62), statistical parity optimizes at 50\% labels (\(\Delta_{SP}=0.16\)), and equal opportunity achieves its best trade-off at 30\% labels (\(\Delta_{EO}=1.57\)).

These results demonstrate that while both models benefit from increased labeled data, EOSP provides \textbf{substantial fairness improvements} at every data scale. Even with minimal labeled sensitive attributes (20\% of individuals), EOSP improves balanced accuracy by 5.7\% while significantly reducing fairness violations. This confirms EOSP's effectiveness across diverse real-world scenarios where labeled sensitive attribute availability may be limited.

\subsection{RQ4: Time Complexity Analysis}

To assess the additional training time complexity introduced by the proposed model-agnostic EOSP method compared to baseline methods, we report the training times for all methods, with and without EOSP, on the German dataset in Table~\ref{tab4}. According to the results in Table~\ref{tab4}, the additional training time introduced by EOSP is less than one second, making the overhead negligible.

\begin{table}[h!]
    \centering
    \caption{Evaluation of Training Time Performance on the German Dataset.}\label{tab4}
    \resizebox{0.7\textwidth}{!}{%
    \begin{tabular}{lrlr}
    \toprule
    \textbf{Method} & \textbf{Time (Seconds)} & \textbf{Method-Fairness} & \textbf{Time (Seconds)} \\
    \midrule
    GCN           & 1.93  & GCN-EOSP      & 2.52  \\
    SAGE          & 1.94  & SAGE-EOSP     & 2.56  \\
    FairGNN       & 1.39  & FairGNN-EOSP  & 1.99  \\
    FairVGNN      & 47.01 & FairVGNN-EOSP & 49.93 \\
    CAF           & 10.30 & CAF-EOSP      & 11.75 \\
    FairGB           & 2.21 & FairGB-EOSP      & 3.07 \\
    DAB-GCN           & 5.32 & DAB-GCN-EOSP      & 6.24 \\
    \bottomrule
    \end{tabular}
    }
\end{table}

\section{Conclusion}\label{sec:conclusion}

Fairness is a central principle in the European guidelines for high-risk AI systems, which emphasize the need for effective methods to avoid bias. However, the accuracy of algorithms is still crucial, and many fairness techniques lead to an undesirable loss of accuracy. In this paper, we present a novel, model-agnostic approach to improve fairness through equal opportunity and statistical parity regularizations. Our approach is applicable to any node classification model. It effectively mitigates bias while providing higher accuracy than conventional fairness methods. Moreover, when applied to a given model the siza of that model remains the same since no additional design is added.

A key advantage of our method is its practical feasibility: it requires only access to sensitive attributes and class labels of labeled nodes during training. This is particularly valuable for real-world applications, where full coverage of sensitive attribute labels is often not feasible due to data collection issues and privacy concerns. Our approach fits well with these constraints and provides a realistic solution to mitigate bias. Moreover, our method is easy to configure, with only two adjustable parameters to balance the trade-off between accuracy and fairness, making it both practical and accessible for various applications.

\section*{Appendix}
\label{appendix}
\subsection*{Proof of Combination of Equal Opportunity and Statistical Parity}

In this section, we reproduce the proof from \cite{defrance2025maximal} demonstrating the simultaneous satisfaction of both Equal Opportunity (EO) and Statistical Parity (SP). For completeness, we formalize these fairness definitions using the confusion matrix variables (Table~\ref{tab:confusion}) for two sensitive attribute groups, $a$ and $b$.

\begin{table}[h]
\centering
\caption{Symbolic representation of a confusion matrix.}
\label{tab:confusion}
\begin{tabular}{|c|c|c|}
\hline
 & \multicolumn{2}{|c|}{\textbf{Predicted}} \\ 
\hline
\textbf{Actual} & \textbf{Positive} & \textbf{Negative} \\ 
\hline
\textbf{Positive} & True Positive (TP) & False Negative (FN) \\ 
\hline
\textbf{Negative} & False Positive (FP) & True Negative (TN) \\ 
\hline
\end{tabular}
\end{table}

We normalize the confusion matrices such that the four entries in each matrix sum to 1, which simplifies subsequent calculations. The confusion matrix values for each sensitive attribute group are constrained by the underlying dataset characteristics:

\begin{equation}
\begin{cases}
TP_{a} + FN_{a} = 1 - x, \\ 
TN_{a} + FP_{a} = x, \\ 
TP_{b} + FN_{b} = 1 - y, \\ 
TN_{b} + FP_{b} = y, \\[6pt]
0 \leq TP_{a} \leq 1 - x, \quad 0 \leq TP_{b} \leq 1 - y, \\ 
0 \leq TN_{a} \leq x, \quad 0 \leq TN_{b} \leq y, \\ 
0 \leq FP_{a} \leq x, \quad 0 \leq FP_{b} \leq y, \\ 
0 \leq FN_{a} \leq 1 - x, \quad 0 \leq FN_{b} \leq 1 - y.
\end{cases}
\label{eq:conf_constraints}
\end{equation}

In Equation~\ref{eq:conf_constraints}, the quantities $1-x$ and $1-y$ correspond to the base rates of groups $a$ and $b$, respectively, where the base rate is defined as the proportion of positive individuals within each group’s population.

To analyze the compatibility of Equal Opportunity and Statistical Parity, we first define what it means for a combination of fairness definitions to be possible, following \cite{defrance2025maximal}.

\begin{definition}[Possibility of Fairness Combinations]
A combination of two fairness definitions is deemed \textbf{possible} if, after combining their constraints, for all pairs of base rates $(1-x, 1-y)$ belonging to a subset of $[0,1]^2$ with non-zero Lebesgue measure, all elements of both confusion matrices can still take values within subsets of $[0,1]$ that also have non-zero Lebesgue measure.
\end{definition}

We now express Equal Opportunity (EO) and Statistical Parity (SP) using the confusion matrix variables from Table~\ref{tab:confusion}.

\textbf{Equal Opportunity (EO)} (Equation~\ref{eq:evalEO}) requires that the True Positive Rate (TPR) is equal across groups:
\begin{equation}
\mathrm{TPR}_a = \mathrm{TPR}_b 
\quad \Longleftrightarrow \quad 
\frac{TP_a}{TP_a + FN_a} = \frac{TP_b}{TP_b + FN_b}.
\label{eq:eo_constraint}
\end{equation}

Using the base rate constraints from Equation~\ref{eq:conf_constraints}, where $TP_a + FN_a = 1-x$ and $TP_b + FN_b = 1-y$, the EO condition simplifies to:
\begin{equation}
\frac{TP_a}{TP_a + FN_a} = \frac{TP_b}{TP_b + FN_b} 
\quad \Longleftrightarrow \quad 
\frac{TP_a}{1-x} = \frac{TP_b}{1-y}
\label{eq:eo_simplified}
\end{equation}

\textbf{Statistical Parity (SP)} (Equation~\ref{eq:evalSP}) requires that the probability of a positive prediction is equal across groups:

\begin{equation}
\frac{TP_a + FP_a}{TP_a + FP_a + TN_a + FN_a} = \frac{TP_b + FP_b}{TP_b + FP_b + TN_b + FN_b}
\end{equation}

Since the confusion matrices are normalized such that $TP_g + FP_g + TN_g + FN_g = 1$ for $g \in \{a, b\}$, this reduces to:
\begin{equation}
TP_a + FP_a = TP_b + FP_b
\label{eq:sp_simplified}
\end{equation}

Combining Equal Opportunity (Equation~\ref{eq:eo_simplified}) and Statistical Parity (Equation~\ref{eq:sp_simplified}) yields:
\begin{equation}
\begin{cases}
FN_a = 1 - x - \dfrac{1-x}{1-y}TP_{b}, \\ 
TN_{a} = x - \dfrac{x-y}{1-y}TP_{b} - FP_{b}, \\ 
FN_{b} = 1 - y - TP_{b}, \\ 
TN_{b} = y - FP_{b},\\
TP_{a} = \dfrac{1-x}{1-y}TP_{b},\\
FP_{a} = \dfrac{x-y}{1-y}TP_{b} + FP_{b}.
\end{cases}
\label{eq:conf_constraints2}
\end{equation}

Combining this system with the inequalities from Equation~\ref{eq:conf_constraints} allows us to compute feasible ranges for the free variables and the base rates:
\begin{equation}
\begin{cases}
\dfrac{1}{1-y}TP_{b} \leq 1, \\
\dfrac{1-x}{1-y}TP_{b} \geq 0, \\
FP_{b} \leq x + \dfrac{-x+y}{1-y} TP_{b}, \\
\dfrac{-x+y}{1-y}TP_{b} \leq FP_{b}, \\
TP_{b} \leq 1-y, \\
TP_{b} \geq 0, \\
FP_{b} \leq y, \\
0 \leq \dfrac{1-x}{1-y}, \\
\dfrac{1}{1-y}TP_{b} \leq 1, \\
\dfrac{y-x}{1-y} \leq \dfrac{FP_{b}}{TP_{b}}, \\
FP_{b} + \dfrac{x-y}{1-y} TP_b \leq x
\end{cases}
\Longleftrightarrow
\begin{cases}
\dfrac{y-x}{1-y} \leq \dfrac{FP_{b}}{TP_{b}}, \\
FP_{b} + \dfrac{x-y}{1-y} TP_{b} \leq x.
\end{cases}
\label{eq:conf_constraints3}
\end{equation}

This system of inequalities shows a specific feasible range for the free variables $FP_{b}$ and $TP_{b}$, while the base rates $1-x$ and $1-y$ remain unconstrained. Therefore, the Lebesgue measure is non-zero for each free variable over a set of base rates with a non-zero two-dimensional Lebesgue measure. Consequently, the combination of equal opportunity and statistical parity is possible.

\begin{thebibliography}{10}

\bibitem{caton2024fairness}
Simon Caton and Christian Haas.
\newblock Fairness in machine learning: A survey.
\newblock {\em ACM Computing Surveys}, 56(7):1--38, 2024.

\bibitem{mehrabi2021survey}
Ninareh Mehrabi, Fred Morstatter, Nripsuta Saxena, Kristina Lerman, and Aram Galstyan.
\newblock A survey on bias and fairness in machine learning.
\newblock {\em ACM computing surveys (CSUR)}, 54(6):1--35, 2021.

\bibitem{madden2013teens}
Mary Madden, Amanda Lenhart, Sandra Cortesi, Urs Gasser, Maeve Duggan, Aaron Smith, and Meredith Beaton.
\newblock Teens, social media, and privacy.
\newblock {\em Pew Research Center}, 21(1055):2--86, 2013.

\bibitem{scarselli2008graph}
Franco Scarselli, Marco Gori, Ah~Chung Tsoi, Markus Hagenbuchner, and Gabriele Monfardini.
\newblock The graph neural network model.
\newblock {\em IEEE transactions on neural networks}, 20(1):61--80, 2008.

\bibitem{fan2019graph}
Wenqi Fan, Yao Ma, Qing Li, Yuan He, Eric Zhao, Jiliang Tang, and Dawei Yin.
\newblock Graph neural networks for social recommendation.
\newblock In {\em The world wide web conference}, pages 417--426, 2019.

\bibitem{wu2022graph}
Shiwen Wu, Fei Sun, Wentao Zhang, Xu~Xie, and Bin Cui.
\newblock Graph neural networks in recommender systems: a survey.
\newblock {\em ACM Computing Surveys}, 55(5):1--37, 2022.

\bibitem{oneto2020fairness}
Luca Oneto and Silvia Chiappa.
\newblock Fairness in machine learning.
\newblock In {\em Recent trends in learning from data: Tutorials from the inns big data and deep learning conference (innsbddl2019)}, pages 155--196. Springer, 2020.

\bibitem{del2020review}
Eustasio Del~Barrio, Paula Gordaliza, and Jean-Michel Loubes.
\newblock Review of mathematical frameworks for fairness in machine learning.
\newblock {\em arXiv preprint arXiv:2005.13755}, 2020.

\bibitem{barocas2023fairness}
Solon Barocas, Moritz Hardt, and Arvind Narayanan.
\newblock {\em Fairness and machine learning: Limitations and opportunities}.
\newblock MIT press, 2023.

\bibitem{wang2022improving}
Yu~Wang, Yuying Zhao, Yushun Dong, Huiyuan Chen, Jundong Li, and Tyler Derr.
\newblock Improving fairness in graph neural networks via mitigating sensitive attribute leakage.
\newblock In {\em Proceedings of the 28th ACM SIGKDD conference on knowledge discovery and data mining}, pages 1938--1948, 2022.

\bibitem{buyl2020debayes}
Maarten Buyl and Tijl De~Bie.
\newblock Debayes: a bayesian method for debiasing network embeddings.
\newblock In {\em International Conference on Machine Learning}, pages 1220--1229. PMLR, 2020.

\bibitem{dai2021say}
Enyan Dai and Suhang Wang.
\newblock Say no to the discrimination: Learning fair graph neural networks with limited sensitive attribute information.
\newblock In {\em Proceedings of the 14th ACM International Conference on Web Search and Data Mining}, pages 680--688, 2021.

\bibitem{dong2022edits}
Yushun Dong, Ninghao Liu, Brian Jalaian, and Jundong Li.
\newblock Edits: Modeling and mitigating data bias for graph neural networks.
\newblock In {\em Proceedings of the ACM Web Conference 2022}, pages 1259--1269, 2022.

\bibitem{rahman2019fairwalk}
Tahleen Rahman, Bartlomiej Surma, Michael Backes, and Yang Zhang.
\newblock Fairwalk: Towards fair graph embedding.
\newblock 2019.

\bibitem{agarwal2021towards}
Chirag Agarwal, Himabindu Lakkaraju, and Marinka Zitnik.
\newblock Towards a unified framework for fair and stable graph representation learning.
\newblock In {\em Uncertainty in Artificial Intelligence}, pages 2114--2124. PMLR, 2021.

\bibitem{Ma_2022}
Jing Ma, Ruocheng Guo, Mengting Wan, Longqi Yang, Aidong Zhang, and Jundong Li.
\newblock Learning fair node representations with graph counterfactual fairness.
\newblock February 2022.

\bibitem{Guo_2023}
Zhimeng Guo, Jialiang Li, Teng Xiao, Yao Ma, and Suhang Wang.
\newblock Towards fair graph neural networks via graph counterfactual.
\newblock October 2023.

\bibitem{hu2024migrate}
YanMing Hu, TianChi Liao, JiaLong Chen, Jing Bian, ZiBin Zheng, and Chuan Chen.
\newblock Migrate demographic group for fair graph neural networks.
\newblock {\em Neural Networks}, 175:106264, 2024.

\bibitem{zhang2024learning}
Guixian Zhang, Debo Cheng, Guan Yuan, and Shichao Zhang.
\newblock Learning fair representations via rebalancing graph structure.
\newblock {\em Information Processing \& Management}, 61(1):103570, 2024.

\bibitem{chen2024fairness}
April Chen, Ryan~A Rossi, Namyong Park, Puja Trivedi, Yu~Wang, Tong Yu, Sungchul Kim, Franck Dernoncourt, and Nesreen~K Ahmed.
\newblock Fairness-aware graph neural networks: A survey.
\newblock {\em ACM Transactions on Knowledge Discovery from Data}, 18(6):1--23, 2024.

\bibitem{spinelli2021fairdrop}
Indro Spinelli, Simone Scardapane, Amir Hussain, and Aurelio Uncini.
\newblock Fairdrop: Biased edge dropout for enhancing fairness in graph representation learning.
\newblock {\em IEEE Transactions on Artificial Intelligence}, 3(3):344--354, 2021.

\bibitem{liu2023generalized}
Zemin Liu, Trung-Kien Nguyen, and Yuan Fang.
\newblock On generalized degree fairness in graph neural networks.
\newblock In {\em Proceedings of the AAAI Conference on Artificial Intelligence}, volume~37, pages 4525--4533, 2023.

\bibitem{kose2023fairness}
O~Deniz Kose, Yanning Shen, and Gonzalo Mateos.
\newblock Fairness-aware graph filter design.
\newblock In {\em 2023 57th Asilomar Conference on Signals, Systems, and Computers}, pages 330--334. IEEE, 2023.

\bibitem{kang2020inform}
Jian Kang, Jingrui He, Ross Maciejewski, and Hanghang Tong.
\newblock Inform: Individual fairness on graph mining.
\newblock In {\em Proceedings of the 26th ACM SIGKDD international conference on knowledge discovery \& data mining}, pages 379--389, 2020.

\bibitem{madras2018learning}
David Madras, Elliot Creager, Toniann Pitassi, and Richard Zemel.
\newblock Learning adversarially fair and transferable representations.
\newblock In {\em International Conference on Machine Learning}, pages 3384--3393. PMLR, 2018.

\bibitem{dwork2012fairness}
Cynthia Dwork, Moritz Hardt, Toniann Pitassi, Omer Reingold, and Richard Zemel.
\newblock Fairness through awareness.
\newblock In {\em Innovations in Theoretical Computer Science Conference}, pages 214--226, 2012.

\bibitem{zafar2017fairness}
Muhammad~Bilal Zafar, Isabel Valera, Manuel~Gomez Rogriguez, and Krishna~P Gummadi.
\newblock Fairness constraints: Mechanisms for fair classification.
\newblock In {\em International Conference on Artificial Intelligence and Statistics}, pages 962--970. PMLR, 2017.

\bibitem{RisserEtAlJMIV2022}
Laurent Risser, Alberto~Gonz\'{a}lez Sanz, Quentin Vincenot, and Jean-Michel Loubes.
\newblock Tackling algorithmic bias in neural-network classifiers using wasserstein-2 regularization.
\newblock {\em J. Math. Imaging Vis.}, 64(6):672–689, 2022.

\bibitem{donini2018empirical}
Michele Donini, Luca Oneto, Shai Ben-David, John~S Shawe-Taylor, and Massimiliano Pontil.
\newblock Empirical risk minimization under fairness constraints.
\newblock {\em Advances in neural information processing systems}, 31, 2018.

\bibitem{gordaliza2019obtaining}
Paula Gordaliza, Eustasio Del~Barrio, Gamboa Fabrice, and Jean-Michel Loubes.
\newblock Obtaining fairness using optimal transport theory.
\newblock In {\em International conference on machine learning}, pages 2357--2365. PMLR, 2019.

\bibitem{venkatasubramanian2021fairness}
Suresh Venkatasubramanian, Carlos Scheidegger, Sorelle Friedler, and Aaron Clauset.
\newblock Fairness in networks: social capital, information access, and interventions.
\newblock In {\em Proceedings of the 27th ACM SIGKDD Conference on Knowledge Discovery \& Data Mining}, pages 4078--4079, 2021.

\bibitem{kang2021fair}
Jian Kang and Hanghang Tong.
\newblock Fair graph mining.
\newblock In {\em Proceedings of the 30th ACM International Conference on Information \& Knowledge Management}, pages 4849--4852, 2021.

\bibitem{kang2022algorithmic}
Jian Kang and Hanghang Tong.
\newblock Algorithmic fairness on graphs: Methods and trends.
\newblock In {\em Proceedings of the 28th ACM SIGKDD Conference on Knowledge Discovery and Data Mining}, pages 4798--4799, 2022.

\bibitem{khajehnejad2022crosswalk}
Ahmad Khajehnejad, Moein Khajehnejad, Mahmoudreza Babaei, Krishna~P Gummadi, Adrian Weller, and Baharan Mirzasoleiman.
\newblock Crosswalk: Fairness-enhanced node representation learning.
\newblock In {\em Proceedings of the AAAI Conference on Artificial Intelligence}, volume~36, pages 11963--11970, 2022.

\bibitem{laclau2021all}
Charlotte Laclau, Ievgen Redko, Manvi Choudhary, and Christine Largeron.
\newblock All of the fairness for edge prediction with optimal transport.
\newblock In {\em International Conference on Artificial Intelligence and Statistics}, pages 1774--1782. PMLR, 2021.

\bibitem{dong2023interpreting}
Yushun Dong, Song Wang, Jing Ma, Ninghao Liu, and Jundong Li.
\newblock Interpreting unfairness in graph neural networks via training node attribution.
\newblock In {\em Proceedings of the AAAI Conference on Artificial Intelligence}, volume~37, pages 7441--7449, 2023.

\bibitem{zhu2024fair}
Yuchang Zhu, Jintang Li, Zibin Zheng, and Liang Chen.
\newblock Fair graph representation learning via sensitive attribute disentanglement.
\newblock In {\em Proceedings of the ACM Web Conference 2024}, pages 1182--1192, 2024.

\bibitem{lee2025disentangling}
Yeon-Chang Lee, Hojung Shin, and Sang-Wook Kim.
\newblock Disentangling, amplifying, and debiasing: Learning disentangled representations for fair graph neural networks.
\newblock In {\em Proceedings of the AAAI Conference on Artificial Intelligence}, volume~39, pages 12013--12021, 2025.

\bibitem{liu5320563rethinking}
Chuxun Liu, Debo Cheng, Qingfeng Chen, Jiuyong Li, Lin Liu, and Rongyao Hu.
\newblock Rethinking fair graph representation learning via structural rebalancing.
\newblock {\em Available at SSRN 5320563}.

\bibitem{li2024rethinking}
Zhixun Li, Yushun Dong, Qiang Liu, and Jeffrey~Xu Yu.
\newblock Rethinking fair graph neural networks from re-balancing.
\newblock In {\em Proceedings of the 30th ACM SIGKDD conference on knowledge discovery and data mining}, pages 1736--1745, 2024.

\bibitem{li2024contrastive}
Chengyu Li, Debo Cheng, Guixian Zhang, and Shichao Zhang.
\newblock Contrastive learning for fair graph representations via counterfactual graph augmentation.
\newblock {\em Knowledge-Based Systems}, 305:112635, 2024.

\bibitem{bondy2008graph}
John~Adrian Bondy and Uppaluri Siva~Ramachandra Murty.
\newblock {\em Graph theory}.
\newblock Springer Publishing Company, Incorporated, 2008.

\bibitem{kipf2016semi}
Thomas~N Kipf and Max Welling.
\newblock Semi-supervised classification with graph convolutional networks.
\newblock {\em arXiv preprint arXiv:1609.02907}, 2016.

\bibitem{xu2019powerful}
Keyulu Xu, Weihua Hu, Jure Leskovec, and Stefanie Jegelka.
\newblock How powerful are graph neural networks?, 2019.

\bibitem{hamilton2017inductive}
Will Hamilton, Zhitao Ying, and Jure Leskovec.
\newblock Inductive representation learning on large graphs.
\newblock {\em Advances in neural information processing systems}, 30, 2017.

\bibitem{liu2022graph}
Zhiyuan Liu and Jie Zhou.
\newblock Graph attention networks.
\newblock In {\em Introduction to graph neural networks}, pages 39--41. Springer, 2022.

\bibitem{chen2023fairness}
April Chen, Ryan~A Rossi, Namyong Park, Puja Trivedi, Yu~Wang, Tong Yu, Sungchul Kim, Franck Dernoncourt, and Nesreen~K Ahmed.
\newblock Fairness-aware graph neural networks: A survey.
\newblock {\em ACM Transactions on Knowledge Discovery from Data}, 2023.

\bibitem{defrance2025maximal}
MaryBeth Defrance and Tijl De~Bie.
\newblock Maximal combinations of fairness definitions.
\newblock {\em Journal of Artificial Intelligence Research}, 82:1495--1579, 2025.

\bibitem{akiba2019optuna}
Takuya Akiba, Shotaro Sano, Toshihiko Yanase, Takeru Ohta, and Masanori Koyama.
\newblock {O}ptuna: A next-generation hyperparameter optimization framework.
\newblock In {\em The 25th ACM SIGKDD International Conference on Knowledge Discovery \& Data Mining}, pages 2623--2631, 2019.

\bibitem{asuncion2007uci}
Arthur Asuncion and David Newman.
\newblock Uci machine learning repository, 2007.

\bibitem{jordan2015effect}
Kareem~L Jordan and Tina~L Freiburger.
\newblock The effect of race/ethnicity on sentencing: Examining sentence type, jail length, and prison length.
\newblock {\em Journal of Ethnicity in Criminal Justice}, 13(3):179--196, 2015.

\bibitem{yeh2009comparisons}
I-Cheng Yeh and Che-hui Lien.
\newblock The comparisons of data mining techniques for the predictive accuracy of probability of default of credit card clients.
\newblock {\em Expert systems with applications}, 36(2):2473--2480, 2009.

\bibitem{takac2012data}
Lubos Takac and Michal Zabovsky.
\newblock Data analysis in public social networks.
\newblock In {\em International scientific conference and international workshop present day trends of innovations}, volume~1, 2012.

\bibitem{dong2023fairness}
Yushun Dong, Jing Ma, Song Wang, Chen Chen, and Jundong Li.
\newblock Fairness in graph mining: A survey.
\newblock {\em IEEE Transactions on Knowledge and Data Engineering}, 35(10):10583--10602, 2023.

\end{thebibliography}
\end{document}